%% file: main.tex
\documentclass[10pt,twocolumn,letterpaper]{article}

\usepackage[pagenumbers]{iccv}      


\definecolor{iccvblue}{rgb}{0.21,0.49,0.74}
\usepackage[pagebackref,breaklinks,colorlinks,allcolors=iccvblue]{hyperref}
\usepackage[accsupp]{axessibility}  
\usepackage{balance}
\usepackage{multirow}
\usepackage{tabularx}
\usepackage{booktabs}
\usepackage{xcolor}
\usepackage{bbding}
\usepackage{pifont}
\usepackage{amsmath}
\usepackage{float}
\usepackage{algorithm}
\usepackage{algorithmic}
\usepackage{makecell}
\def\paperID{15507} 
\def\confName{ICCV}
\def\confYear{2025}

\title{mmCooper: A Multi-agent Multi-stage Communication-efficient and Collaboration-robust Cooperative Perception Framework}

\author{Bingyi Liu\textsuperscript{1}, Jian Teng\textsuperscript{1}, Hongfei Xue\textsuperscript{2*}, Enshu Wang\textsuperscript{3}\thanks{Corresponding authors. Our code is available at \url{https://github.com/tengjjj/mmCooper}.}, Chuanhui Zhu\textsuperscript{1}, Pu Wang\textsuperscript{2}, Libing Wu\textsuperscript{3}\\
\textsuperscript{1}Wuhan University Of Technology, \textsuperscript{2}University of North Carolina at Charlotte, \textsuperscript{3}Wuhan University\\
{\tt\small \{byliu, tengjian, zhuchuanhui\}@whut.edu.cn}, \\
{\tt\small \{hongfei.xue, Pu.Wang\}@charlotte.edu},
{\tt\small \{wanges17, wu\}@whu.edu.cn}
}

\begin{document}
\maketitle
\input{sec/0_abstract}    
\input{sec/1_intro}

\input{sec/2_related_works.tex}

\input{sec/3_methodology.tex}
\input{sec/4_experiments.tex}

\input{sec/5_conclusion.tex}
\input{sec/supp}

{
    \small
    \clearpage
    \bibliographystyle{ieeenat_fullname}
    \bibliography{main}
}
\end{document}

%% file: sec/0_abstract.tex
\begin{abstract}
Collaborative perception significantly enhances individual vehicle perception performance through the exchange of sensory information among agents.
However, real-world deployment faces challenges due to bandwidth constraints and inevitable calibration errors during information exchange.
To address these issues, we propose mmCooper, a novel multi-agent, multi-stage, communication-efficient, and collaboration-robust cooperative perception framework.
Our framework leverages a multi-stage collaboration strategy that dynamically and adaptively balances intermediate- and late-stage information to share among agents, enhancing perceptual performance while maintaining communication efficiency.
To support robust collaboration despite potential misalignments and calibration errors, our framework prevents misleading low-confidence sensing information from transmission and refines the received detection results from collaborators to improve accuracy.
The extensive evaluation results on both real-world and simulated datasets demonstrate the effectiveness of the mmCooper framework and its components.
\end{abstract}

%% file: sec/1_intro.tex
\vspace{-12pt}
\section{Introduction}
\vspace{-5pt}
\label{sec:intro}

With the advancement of autonomous driving systems~\cite{yurtsever2020survey, levinson2011towards}, perception devices such as cameras and LiDARs have been widely deployed.
Although perception technologies have witnessed rapid advancements driven by deep learning-based algorithms, the traditional single-vehicle perception paradigms~\cite{li2022bevformer,liu2023bevfusion,yang2023bevformer} cannot meet the safety and reliability requirements of autonomous vehicles due to unavoidable factors such as object occlusion and limitations in detection range~\cite{zhang2021emp,allig2019alignment,hua2019hierarchical,chen2023dynamic,liu2020vision}.


Benefiting from infrastructure improvements and the advancement of Internet of Vehicles (IoVs) technologies like V2X~\cite{liu2024multi,liu2023efficient}, autonomous vehicles can achieve multi-vehicle collaborative perception by sharing perception information~\cite{wei2024asynchrony,chang2023bev}, which significantly enhances the performance of perception systems.
Although cooperative perception has made progress recently, its practical deployment still faces challenges such as communication resource constraints~\cite{hu2022where2comm,chen2023transiff}, localization errors~\cite{lu2023robust,ni2024self}, and low information fusion efficiency~\cite{yang2023what2comm,wang2020v2vnet}.

In cooperative perception systems, the effectiveness of fusion strategies significantly affects both perception performance and communication efficiency. 
The fusion strategies can be divided into three categories: early, intermediate, and late fusion methods. 
Specifically, early fusion methods~\cite{chen2019cooper,zhang2021emp} aggregate raw sensor observations from all agents, providing abundant information
but requiring immense bandwidth, making this approach costly for communication.
Intermediate fusion methods~\cite{hu2022where2comm,chen2023transiff,yang2023what2comm, yang2023spatio,yang2024how2comm} partially mitigate this issue by sharing encoded features rather than raw data to reduce communication overhead.
However, transmitting the feature map of the entire sensing scenario still results in substantial bandwidth consumption~\cite{hu2022where2comm}.
Late fusion methods~\cite{shi2022vips,song2023cooperative} on the other hand, minimize communication demands by sharing lightweight perception results among agents.
However, this approach is highly sensitive to environmental noise, as even minor disturbances can degrade collaborative quality~\cite{lu2023robust}.
Overall, these above single-stage cooperative perception approaches face challenges in managing the trade-off between perception accuracy and communication efficiency, raising a compelling question: \textit{Can we improve the perception performance while preserving communication efficiency by strategically sharing information across multiple stages to harness the strengths of each fusion stage?}

In addition, the data exchanged by agents inevitably experiences misalignment and contains calibration errors during the synchronization process due to communication delays and pose noise~\cite{huang2023v2x, su20233d}.
These calibration errors can produce misleading features and proposals in subsequent steps, thereby affecting the performance of the perception system. While existing intermediate fusion methods attempt to address the calibration errors through position-wise feature fusion~\cite{li2021learning, li2023learning, liu2020when2com}, they often overlook the critical role of neighboring region information for enhancing collaborative robustness.
Meanwhile, late fusion methods~\cite{cai2023consensus, piazzoni2022copem, su2024collaborative} simply merge bounding boxes from collaborative agents, failing to leverage information-rich intermediate features from the ego agent, which could help refine and calibrate these bounding boxes for higher accuracy.



To this end, we propose \textbf{mmCooper}, a novel \textbf{m}ulti-agent, \textbf{m}ulti-stage, communication-efficient, and collaboration-robust \textbf{cooper}ative perception framework.
The proposed mmCooper framework facilitates effective information aggregation across fusion stages and enhances communication efficiency through carefully designed core components.
Specifically, we design an adaptive multi-stage data fusion mechanism that facilitates multi-agent cooperation by dynamically fusing information at both intermediate and late stages.
Guided by a confidence-based filtering strategy, this mechanism automatically assesses and determines the data volume to be processed at intermediate and late stages.
High-confidence bounding boxes are shared directly, while intermediate features are selectively shared instead of bounding boxes for regions requiring further contextual reference, and low-confidence areas are excluded to prevent propagating misleading information.
This dual-stage fusion approach enhances perception performance while maintaining communication efficiency.

Besides, to further address potential misalignment and calibration noise among agents, we incorporate specific designs in the proposed mmCooper framework for both intermediate and late stages.
In the intermediate stage, we introduce a Multi-scale Offset-aware Fusion Module that not only fuses data from target locations related to cooperative agents’ views but also from nearby regions, adding redundancy to mitigate calibration noise. 
For the late stage, we design a Bounding Box Filtering \& Calibration Module that uses the information-rich intermediate-stage features from the ego agent to filter inaccurate bounding boxes received from other agents and refine the remaining noisy ones.
By integrating these corrected and ego-generated bounding boxes, our mmCooper framework can achieve high perception accuracy and robustness.
To summarize, our main contributions are summarized as follows:
\begin{itemize}
    \item We propose mmCooper, a multi-agent multi-stage communication-efficient collaboration-robust cooperative perception framework. To our knowledge, this is the first framework to address the tradeoff between limited communication bandwidth and desired perception performance by sharing both intermediate and late-stage information for multi-agent collaborative perception.

    \item To address the potential misalignment and calibration error among agents, we design a Multi-scale Offset-aware Fusion Module to integrate spatially adjacent contextual information at different feature scales in the intermediate stage and a Bounding Box Filtering \& Calibration Module to filter and improve the bounding boxes from other agents in the late stage, guided by the information-rich intermediate features.
    
    

    \item 
    Extensive experiments show excellent performance on both real-world and simulated datasets, with a 7.29\%/1.31\%/2.09\% improvement in AP@0.7 on OPV2V~\cite{xu2022opv2v}, DAIR-V2X~\cite{yu2022dair} and V2XSet~\cite{xu2022v2x} datasets over the second-best SOTA methods. Meanwhile, our approach requires only 1/9153, 1/156, and 1/18305 of the communication volume used by comparable methods. Furthermore, our method consistently outperforms previous approaches under various configurations.
\end{itemize}

%% file: sec/2_related_works.tex
\vspace{-5pt}
\section{Related Work}
\label{sec:related_works}
\vspace{-3pt}


\subsection{Multi-Agent Communication}
\vspace{-5pt}
Communication strategies in multi-agent systems have been widely studied~\cite{singh2018learning}. Early works~\cite{tan1993multi,qureshi2008smart, li2010auction} often used predefined protocols or heuristic methods to determine how agents communicate with each other. However, these methods are difficult to generalize to complex tasks. Recently, several learning-based communication strategies have been proposed to adapt to diverse scenario applications. For instance, Vain~\cite{hoshen2017vain} utilizes attention neural structures to specify the information that needs to be shared in agent interactions. ATOC~\cite{jiang2018learning} introduces recurrent units to decide who agents communicate with by receiving local observations and action intentions from other agents. TarMAC~\cite{das2019tarmac} designs an architecture oriented towards reinforcement learning, learning to communicate from task-specific rewards. CommNet~\cite{sukhbaatar2016learning} learns continuous communication in multi-agent systems. Most of these previous works have considered decision-making tasks and have employed reinforcement learning due to the lack of explicit supervision.
Different from these works, our paper focuses on LiDAR-based 3D perception tasks in complex autonomous driving scenarios and achieves efficient communication between agents using a multi-stage communication strategy.

\vspace{-3pt}
\subsection{Collaborative Perception}
\vspace{-3pt}

Collaborative perception addresses the limitations of single-agent perception by aggregating perception information from surrounding agents to adapt to complex and dynamic autonomous driving environments. Current work is categorized into early, intermediate, and late fusion based on collaboration stages. Early fusion~\cite{chen2019cooper,zhang2023robust,zhang2021emp} shares raw point cloud data among agents, achieving high perception performance but consuming significant bandwidth resources. EMP~\cite{zhang2021emp} achieved scalability and adaptability on highly fluctuating open roads by dynamically partitioning the point cloud range shared by agents. Late fusion~\cite{shi2022vips,xu2023model,song2023cooperative} shares lightweight proposals between agents but performs poorly and is sensitive to noise. VIPS~\cite{shi2022vips} implemented efficient graph-structured matching to enable object-level fusion. Intermediate fusion~\cite{liu2020when2com,hu2022where2comm,xu2022v2x,zhang2024ermvp,li2021learning} shared features among agents, yet complete Intermediate features still incur high bandwidth consumption. Recent work focused on novel communication mechanisms to filter Intermediate features and reduce communication volume. for instance, Where2comm~\cite{hu2022where2comm} employed a confidence-based scheme to guide agents in focusing on sharing spatially critical information. ERMVP~\cite{zhang2024ermvp} adopted a hierarchical feature sampling strategy to reduce communication overhead while leveraging sparse consensus features to mitigate localization errors.
Different from these works, our framework conducts multi-stage fusion instead of single-stage which balances the perception performance and communication overhead.
As far as we know, only a few works have considered multi-stage settings.
Zhang \textit{et al.}~\cite{zhang2024data} proposes a defense method that combines raw data and intermediate features to fuse redundant information, thereby reducing the impact of spurious data.
However, this work focus on security issues related to collaborative perception instead of improving sensing performance.
Xie \textit{et al.}~\cite{xie2022soft} applies reinforcement learning for dynamic partitioning of early, intermediate, and late-stage information. However, this work merely divides the data without integrating information across stages.
In contrast to these works, this paper introduces a collaborative perception framework that aims to improve sensing performance by conducting multi-stage data fusion.

\begin{figure}[t]
    \begin{center}
    \vspace{-5pt}
    \includegraphics[width=1.0\linewidth]{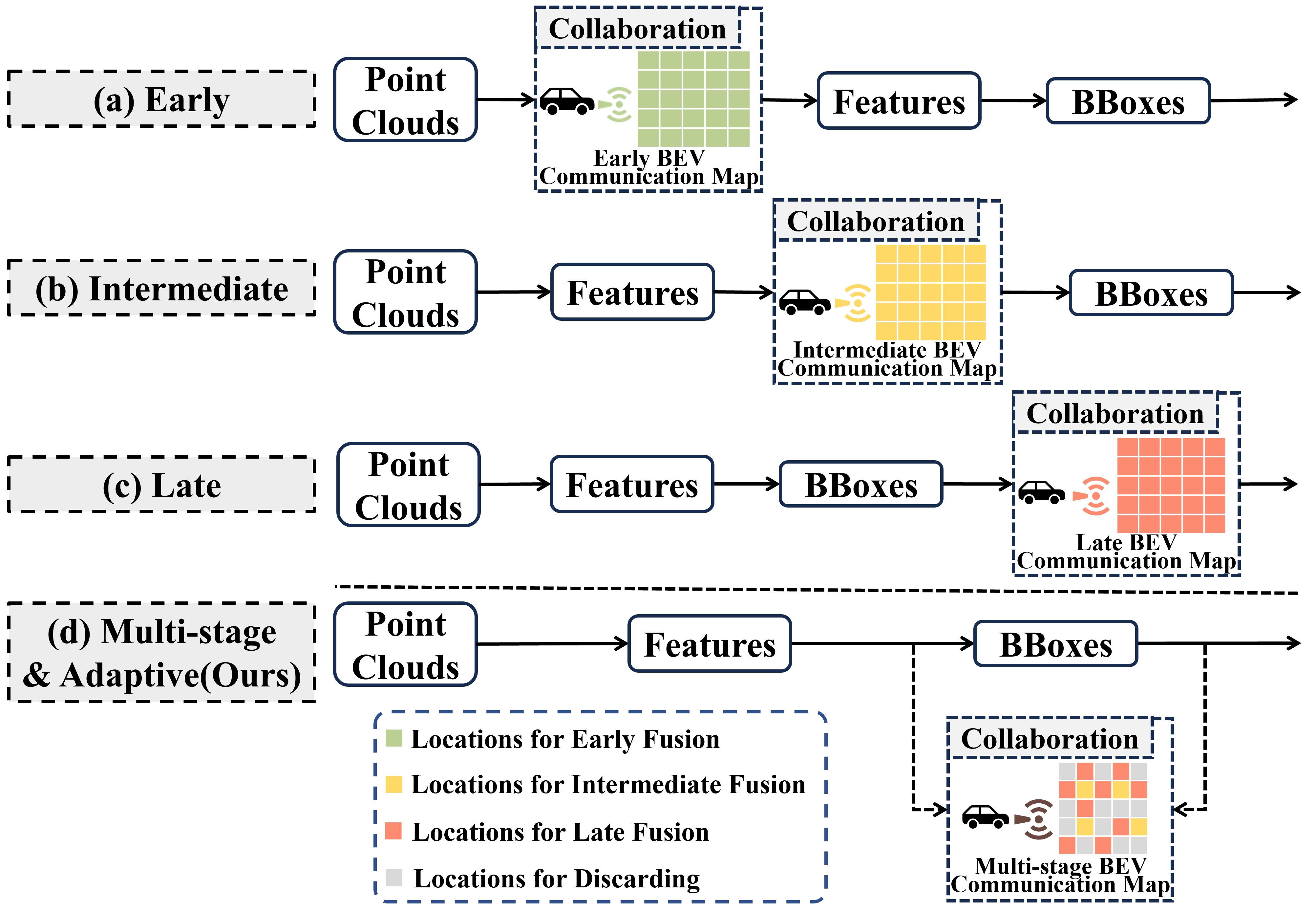}
    \vspace{-23pt}
    \caption{\footnotesize{Comparison between our proposed multi-stage mmCooper framework and existing methods. The BEV (Bird’s Eye View) communication map illustrates the information shared by agents at each scene location. (a)(b)(c) depict existing methods, which transmit the entire scene’s data in a single stage without considering sensing confidence, leading to excessive communication overhead and degraded performance. (d) demonstrates how mmCooper selectively transmits non-overlapping information across multiple stages, reducing communication costs and enhancing model performance.}}
    \label{fig:commuication}
    \vspace{-29pt}
    \end{center}
\end{figure}

%% file: sec/3_methodology.tex
\vspace{-3pt}
\section{Methodology}
\label{sec:methodology}
\vspace{-3pt}
\subsection{Overview}
\vspace{-3pt}

\begin{figure*}
    \centering
    \includegraphics[width=0.92\linewidth]{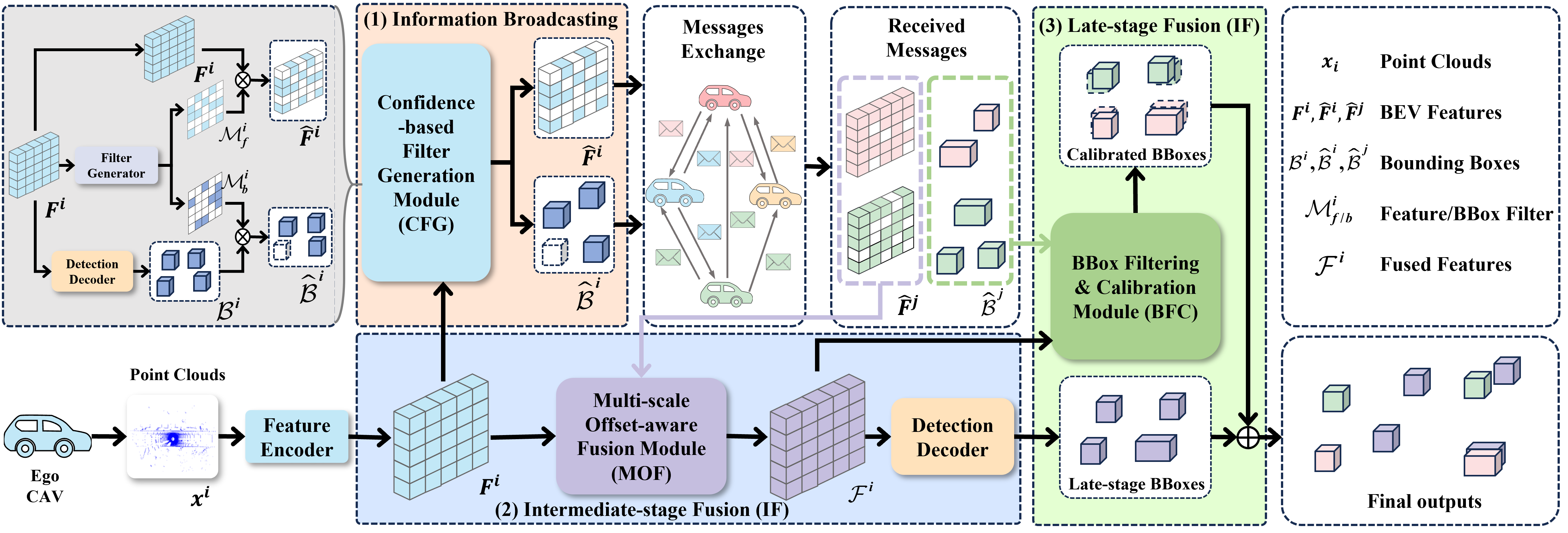}
    \vspace{-13pt}
    \caption{\footnotesize Overview of mmCooper framework: a cooperative perception system with adaptive multi-stage fusion. It consists of three main components: 
    (1) Information Broadcasting (\cref{subsec:broadcasting}) filters features and bounding boxes to achieve bandwidth efficiency; (2) Intermediate-stage Fusion (\cref{subsec:intermdiate fusion}) captures surrounding information from the received feature maps for robust feature fusion; (3) Late-stage Fusion (\cref{subsec:late}) utilizes the information-rich fused features for Filtering \& Calibration of the received bounding boxes.}
    \label{fig:framework}
    \vspace{-18pt}
\end{figure*}

As illustrated in \cref{fig:commuication}, our key idea is to conduct selective multi-stage agent collaboration, rather than the single-stage fusion used in existing methods.
Conventional approaches indiscriminately transmit data encompassing the entire scene through a single stage, introducing undesired low-quality sensing information during communication.
This not only increases communication overhead but also complicates fusion, as the system must first evaluate the quality of received data to ensure reliable results.
In contrast, our approach adopts a selective multi-stage collaboration strategy, where the transmission decision, whether to send data and at which stage, is determined based on the sensing confidence level of each location.
For high-confidence regions, we transmit late-stage bounding boxes, minimizing communication costs. For moderate-confidence regions, we transmit intermediate-stage features for richer information. For low-confidence regions, transmission is suppressed to prevent misleading other agents.
Additionally, our framework incorporates alignment and calibration corrections during both intermediate and late-stage fusion, further enhancing the reliability of the fused results.
The details of our proposed mmCooper framework are illustrated in \cref{fig:framework}, which comprises three main components: 
(1) the Information Broadcasting, (2) the Intermediate-stage Fusion, and  (3) the Late-stage Fusion.
The encoded point cloud features will be sent to two branches.
In one branch, the Information Broadcasting (\cref{subsec:broadcasting}) employs a Confidence-based Filter Generation Module to dynamically determine whether data should be suppressed, transmitted at the intermediate stage, or transmitted at the late stage.
In the other branch, the Intermediate-stage Fusion (\cref{subsec:intermdiate fusion}) contains the Multi-scale Offset-aware Fusion Module, which uses cross-attention to integrate features from both target and neighboring regions.
This module mitigates potential misalignment between features from the ego and other agents, ensuring robust feature fusion.
In Late-stage Fusion (\cref{subsec:late}), a BBox Filtering \& Calibration Module removes inaccurate bounding boxes from other agents and refines the remaining ones using the ego agent’s information-rich intermediate features. 
The calibrated bounding boxes from all agents will be merged with those generated by the ego agent to produce the final outputs.


\subsection{Observation Encoding}
\vspace{-3pt}
\label{subsec:observation}

In a multi-agent cooperative perception scenario, consider the collaboration of $n$ agents, represented by the agent set $N=\left \{ 1,...,n \right \}$. The $i$-th agent serves as the ego vehicle and the remaining $n-1$ agents act as collaborators.
All the agents first broadcast basic positional information to their collaborators, including coordinates and heading angle.
In this way, the agents obtain the necessary information to project the information from the collaborators' systems to the ego vehicle's coordinate system. 
The ego agent processes its point cloud through a shared encoder to extract Bird’s Eye View (BEV) features, represented as \(F_{}^{i} = \psi_{E}\left ( x_{}^{i} \right ) \in \mathbb{R}^{C\times H\times W}\), where \(\psi_{E}\) denotes the shared PointPillar~\cite{lang2019pointpillars} encoder, \( x_{}^{i} \) represents the projected local observation, and \(C\), \(H\), and \(W\) correspond to the channel, height, and width of the feature map, respectively.

\vspace{-2pt}
\subsection{Information Broadcasting}
\vspace{-3pt}



\begin{figure}
    \begin{center}
    \includegraphics[width=1.0\linewidth]{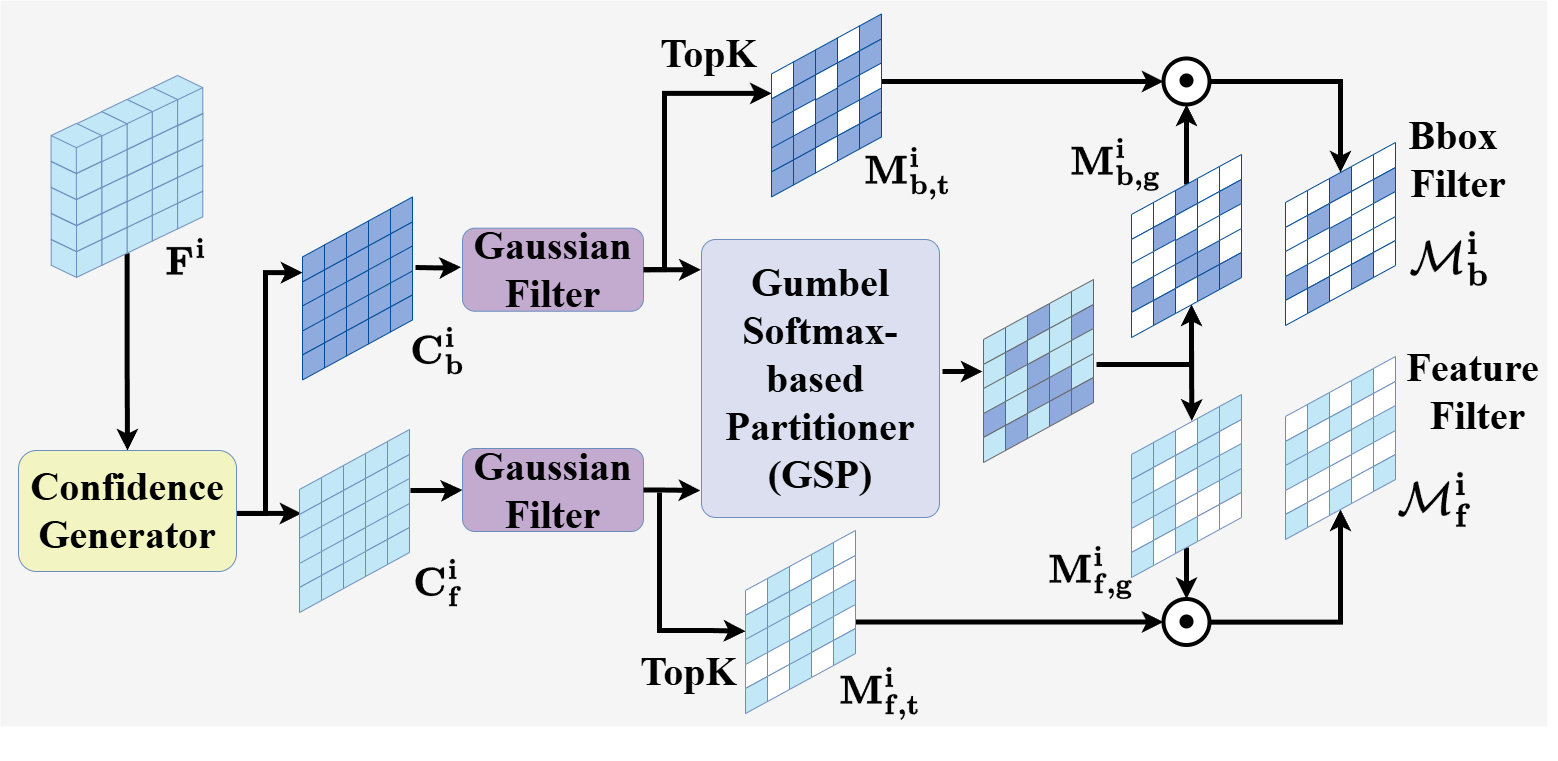}
    \vspace{-23pt}
    \caption{\footnotesize{The design of Filter Generator in CFG Module, generating confidence scores to guide information transmission at each location.
    }}
    \label{fig:broadcasting}    
    \vspace{-25pt}
    \end{center}
\end{figure}

As shown in the top-left corner of \cref{fig:framework}, when conducting information broadcasting, the Confidence-based Filter Generation (CFG) Module dynamically decides whether data at each location should be suppressed, transmitted as intermediate-stage features, or transmitted as late-stage bounding boxes.
Notably, no additional model or data compression techniques~\cite{Ma_2024_WACV} are applied in this framework, only raw features or bounding boxes are transmitted. 
The bounding boxes are generated by the Detection Decoder from the features and the transmitting decision is guided by the confidence filter generated by the Filter Generator (\cref{fig:broadcasting}) based on the sensing confidence, as detailed below.

\label{subsec:broadcasting}

\noindent \textbf{Filter Generator.} 
The encoded BEV features (i.e., $F^{i}$) are fed into this Confidence Generator to produce intermediate-stage confidence map $C_{f}^{i}$ and late-stage confidence map $C_{b}^{i}$ for all the locations of the scene using CNN.
A Gaussian Filter is then applied to the generated confidence maps to smooth out noise and reduce anomalies, improving the selection of critical regions. 
To suppress the information of the locations with low confidence, only the top $p\%$ highest-confidence positions are selected from each confidence map to generate the spatial filters $M_{f/b, t}^{i}$ (i.e., $M_{f, t}^{i}$ or $M_{b, t}^{i}$).
Then, for each location, a classification is required to decide which stage to transmit the data.
However, using the traditional Softmax will prevent the back-propagation of the gradient due to its non-differentiability as a discrete operation.
Instead, we utilize the Gumbel Softmax~\cite{jang2016categorical} to obtain approximate samples from a discrete distribution ensuring the propagation of gradient through the straight-through estimator~\cite{bengio2013estimating} while preserving standard forward propagation.
The Gumbel filter $M_{f/b,g}^{i}$ obtained in this process is given by the following equation: 
\setlength{\abovedisplayskip}{2pt}
\setlength{\belowdisplayskip}{2pt}
\[
M_{f/b,g}^{i}= f_{max}  \left(\frac{exp\left (( logC_{f/b}^{i} + g_{f/b} )/\tau\right ) }{ { {\textstyle \sum_{s}} exp\left (( logC_{s}^{i} + g_{s} )/\tau\right )}  } \right), 
\]
where $g_{f/b}$ is Gumbel noise~\cite{maddison2016concrete}, $\tau$ denotes the temperature parameter of the distribution $(
\tau>0 ) $, and $ s\in {\left \{  f,b\right \} }$.
Based on the above process, our final filter $ \mathcal{M}_{f/b}^{i}$ is represented as:
\[
\mathcal{M}_{f/b}^{i} = M_{f/b,g}^{i} \odot M_{f/b,t}^{i}.
\]
Based on the aforementioned filter, the intermediate-stage feature \( F_{}^{i} \) and the coarse bounding box \( \mathcal{B}_{}^{i} \) are filtered to obtain the filted feature \( \hat{F}_{}^{i} \) and  coarse bounding box \( \hat{\mathcal{B}}_{}^{i} \).


\begin{figure}
    \begin{center}    
    \includegraphics[width=1.0\linewidth]{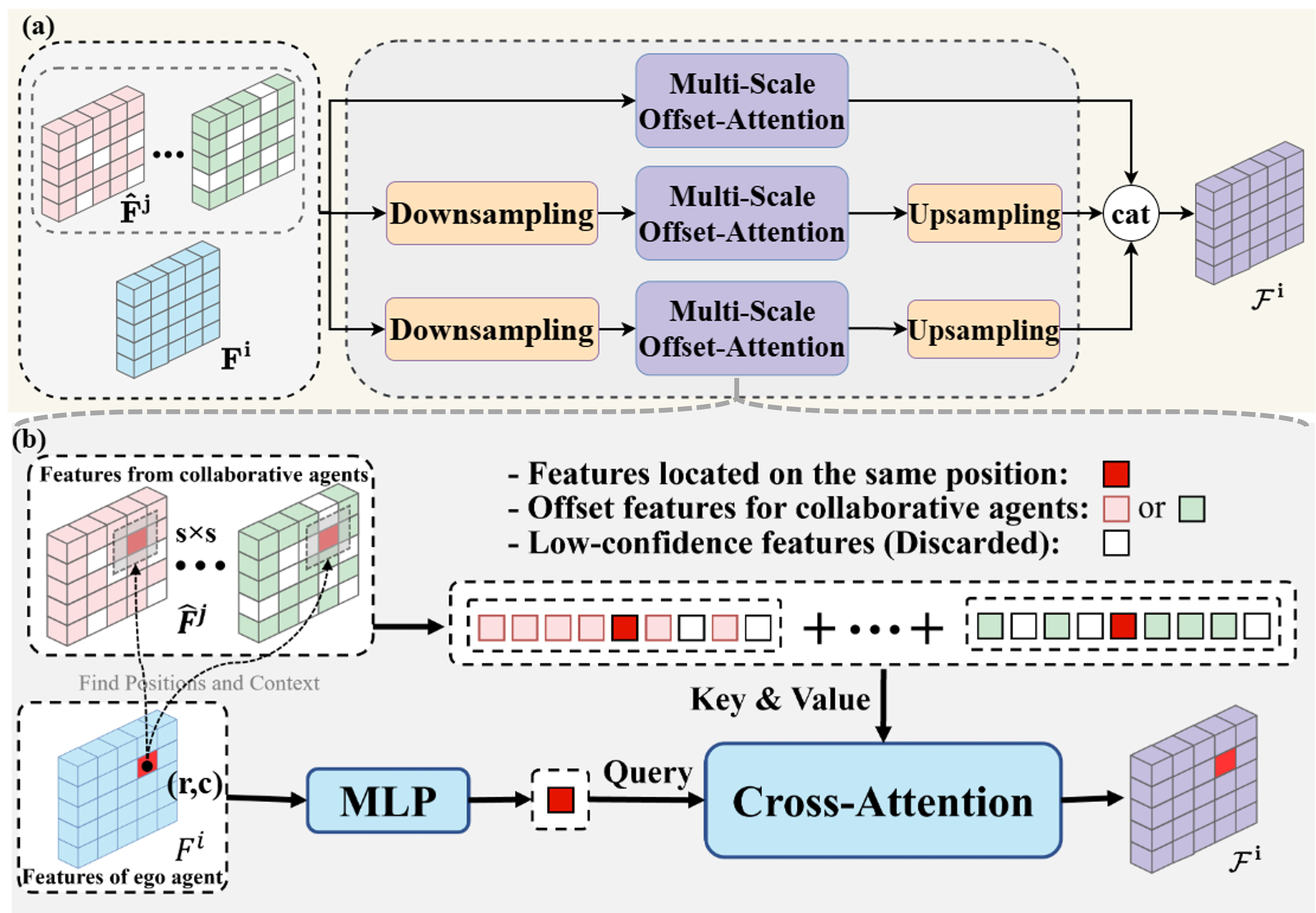}
    \vspace{-20pt}
    \caption{\footnotesize{(a) The Multi-scale Offset-aware Fusion Module. (b) The Multi-scale Offset-aware Attention Module.}}
    \label{fig:intermediate}   
    \vspace{-20pt}
    \end{center}
\end{figure}

\vspace{-2pt}
\subsection{Intermediate-stage Fusion}
\vspace{-3pt}
\label{subsec:intermdiate fusion}

The data shared between agents inevitably contains calibration errors, causing potential misalignment in feature maps.
To address this, we propose a Multi-scale Offset-aware Fusion (MOF) Module in the intermediate-stage fusion, which aggregates information from neighboring regions, allowing the fusion process at each location to account for potential misalignment.
By incorporating the surrounding context, it effectively mitigates the impact of calibration errors, ensuring more accurate and robust agent collaboration.

\begin{figure}
    \begin{center}
    \includegraphics[width=1\linewidth]{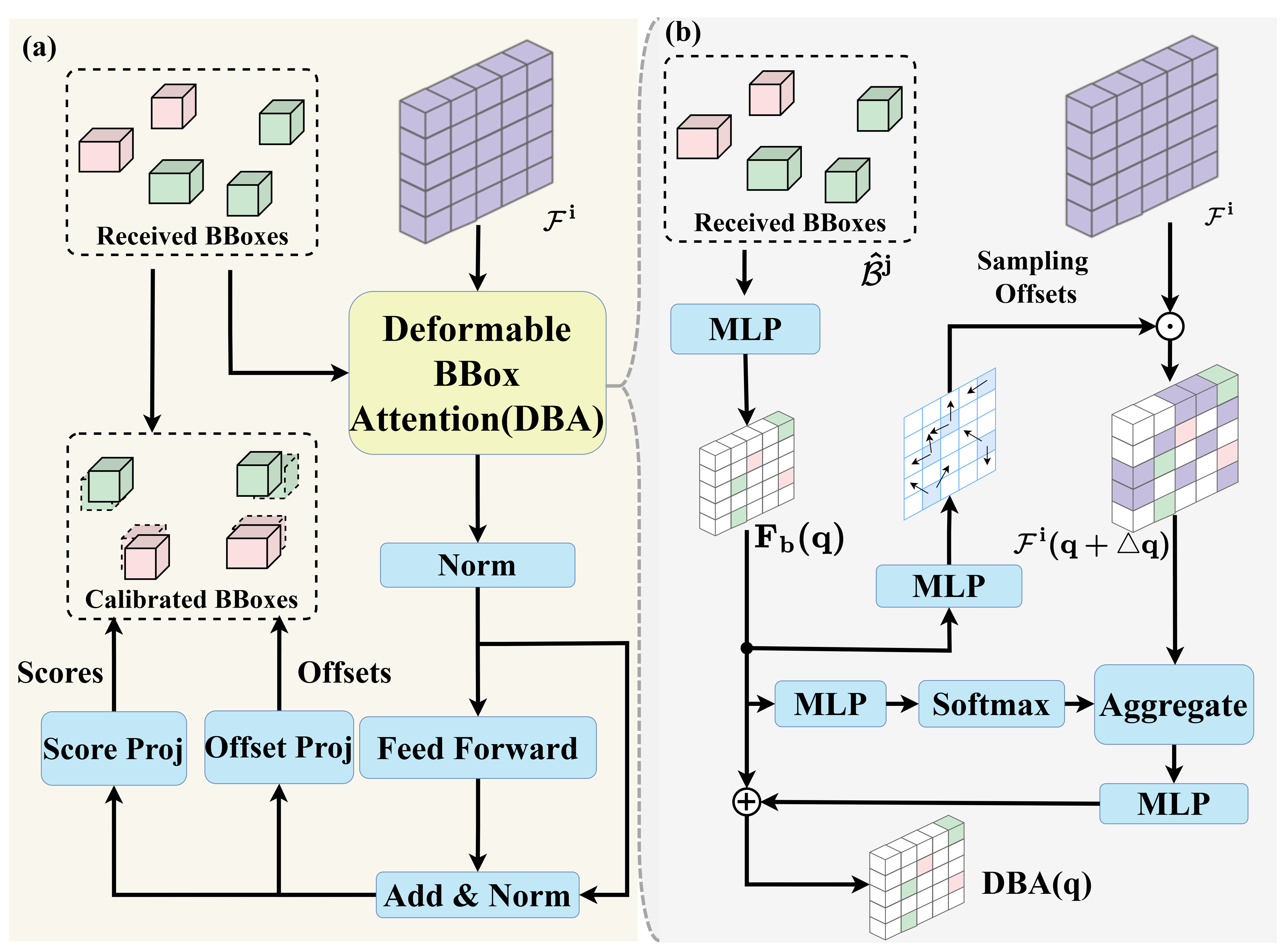}
    \vspace{-20pt}
    \caption{\footnotesize (a) The BBox Filtering \& Calibration (BFC) Module. (b) The Deformable BBox Attention (DBA) Module.}
    \label{fig:late}    
    \vspace{-20pt}
    \end{center}
\end{figure}

\noindent \textbf{Multi-scale Offset-aware Fusion (MOF).}
The overall pipeline of this module is shown in \cref{fig:intermediate}(a). 
First, the ego features and received collaborator features are encoded at three different scales to capture target information of varying sizes.
Next, Multi-scale Offset-aware Attention (MOA) is applied for feature fusion.
The features are then upsampled to a uniform size and concatenated across scales to generate the final fused feature $F^i$.

\noindent \textbf{Multi-scale Offset-aware Attention (MOA).}
As shown in \cref{fig:intermediate}(b), we encode the ego feature at each location into queries using MLP.
Meanwhile, features from a small neighborhood around the corresponding positions in each collaborator’s feature map are extracted as keys and values for cross-attention.
Note that low-confidence features are suppressed by collaborators and not transmitted, making them irrelevant during cross-attention.
Specifically, we denote the focused location as \( (r, c) \) on the feature map, with a neighborhood size of \( s \times s \). The output representation of feature fusion at location \( (r, c) \) is expressed as:
\setlength{\abovedisplayskip}{2pt}
\setlength{\belowdisplayskip}{2pt}
\begin{equation*}
 \mathcal{F}_{r,c}^{i} =  CrossAttn(MLP( F_{r,c}^{i}), \hat{F}_{r,c, s\times s}^{j}, \hat{F}_{r,c,s\times s}^{j}),   
\end{equation*}
Where $\hat{F}_{r,c,s\times s}^{j}$ represents the transmitted collaborator features within an $s\times s$ neighborhood centered at position $(r,c)$.
By selectively aggregating the features at this position and its neighborhood, we achieve collaboratively robust feature fusion.

\subsection{Late-stage Fusion}
\label{subsec:late}

The bounding boxes received from collaborators are often redundant, imprecise, misaligned, or even incorrect.
To address this, the BBox Filtering \& Calibration (BFC) Module in the Late-stage Fusion leverages the information-abundant fused intermediate features to refine the received bounding boxes, either filtering out erroneous ones or correcting their positions for improved accuracy.

\noindent \textbf{BBox Filtering \& Calibration (BFC).}
As depicted in \cref{fig:late}(a), first, the received bounding boxes and the fused intermediate features will be put into the Deformable BBox Attention (DBA) Module as inputs.
It then extracts relevant information for each received bounding box from collaborators, which is processed by a vanilla transformer~\cite{vaswani2017attention} to generate quality scores and positional offsets.
Note that, to prevent unreasonable offsets and scores, we bound the output range using the $Tanh$ activation function.
Finally, bounding boxes with low quality scores are discarded, while the remaining ones are refined by applying the predicted positional offsets. 
The refined bounding boxes are then merged with those detected by the ego agent, followed by Non-Maximum Suppression (NMS) to generate the final results.
To train the BFC module, we compute the offset loss \(\mathcal L_{off}\) using the smooth absolute error loss~\cite{girshick2015fast} and the score loss \(\mathcal L_{score}\) using focal loss~\cite{lin2017focal}.

\noindent \textbf{Deformable BBox Attention (DBA).}
As shown in \cref{fig:late}(b), we first encode the received bounding boxes $\hat{\mathcal{B}}_{}^{j}$ by applying am MLP to their coordinates, size, and orientation information to generate feature representations.
These features are then mapped onto the BEV space based on the bounding box locations, forming the bounding box feature map $F_{b}(q)$.
To enhance these representations, we employ the deformable cross-attention mechanism to selectively attend to and integrate relevant information from the fused intermediate features. 
Specifically, the initial bounding box features $F_{b}(q)$ serve as the query embedding to calculate the offsets of reference points using MLP.
Using these offsets, we retrieve the corresponding reference features from the fused intermediate feature map as $F^i_r=\mathcal{F}_{}^{i}(q+\bigtriangleup q_{m})$.
In the meanwhile, an MLP and Softmax are applied on $F_{b}(q)$ to generate the aggregation weights for $F^i_r$, which are then used to weight and integrate the reference features.
Finally, after another MLP layer, we obtain the enhanced bounding box feature map:
\[
DBA(q)=\sum_{\alpha =1}^{A} W_{\alpha }[\sum_{n=1}^{N} \sum_{m=1}^{M} \omega (W_{\beta }F_{b}(q))F^i_r]+F_{b}(q) ,   
\]
where $A$ is the number of attention heads, $ W_{\alpha} $ and $ W_{\beta} $ are learnable parameters, $m$ represents the number of reference locations, and $\omega$ denotes the softmax operation.

\subsection{Loss Functions}

We use the regression head and classification head of the Detection Decoder~\cite{carion2020end} to generate the bounding box detection results.
The regression results represent the position, size, and yaw angle of each predefined box, expressed as \(O_{reg} = f_{dec}^{r}(\mathcal{F}_{}^{i}) \in \mathbb{R}^{7 \times H \times W}\).
The classification results indicate the confidence score for each predefined box, represented as \(O_{cls} = f_{dec}^{c}(\mathcal{F}_{}^{i}) \in \mathbb{R}^{2 \times H \times W}\). To optimize the proposed system, we use smooth absolute error loss to supervise regression and focal loss for classification, denoted as \(\mathcal L_{reg}\) and \(\mathcal L_{cls}\), respectively.
Combining the previous losses, our total loss is represented as: \[
\mathcal L_{total} = \mathcal L_{reg} + \mathcal L_{cls} + \mathcal L_{off} + \mathcal L_{score} .   
\]

%% file: sec/4_experiments.tex
\section{Experiments}
\label{sec:experiments}

\subsection{Datasets and experimental settings}

\textbf{Datasets.}
We conduct an extensive evaluation on three benchmark datasets, namely OPV2V~\cite{xu2022opv2v},  DAIR-V2X~\cite{yu2022dair}, and V2XSet~\cite{xu2022v2x}.
OPV2V is a large-scale public V2V cooperative perception simulation dataset,
which comprises 73 different scenes, 11,464 frames of point clouds and RGB images, and over 230,000 annotated 3D detection boxes.
The training, validation, and test sets are divided into 6,374, 1,980, and 2,170 frames, respectively.
DAIR-V2X is a large-scale real-world 3D object detection dataset,
including 71,254 frames of point cloud and image data, with the training, validation, and test sets split in a 5:2:3 ratio.
The V2XSet dataset is a large-scale synthetic dataset designed for V2X perception.
It contains a total of 11,447 frames,
which is divided into training, validation, and test sets, consisting of 6,694, 1,920, and 2,833 frames, respectively.

\noindent \textbf{Evaluation metrics.}
To evaluate the 3D object detection performance of the baseline and the proposed framework, we use the average precision (AP)~\cite{everingham2010pascal} at Intersection-over-Union (IoU) thresholds of 0.5 and 0.7 as the evaluation metric.
The communication volume between agents is expressed by the following equation~\cite{hu2022where2comm}:
\[
\textbf{V} =\log_{2}{((\left | M \right |\times H\times W\times C + N_{b}\times 7 )\times 32/8)},
\]
where $\left | M \right |$ represents the feature retention ratio, $N_{b}$ denotes the number of bounding boxes per agent, and 7 is used to describe the coordinates, dimensions, and angle information of each bounding box. 
Each value is represented using float32, with a division by 8 for bytes.

\noindent \textbf{Implementation details.}
We implement our model on the PyTorch toolbox~\cite{paszke2019pytorch} and train it on NVIDIA GeForce RTX 4090 GPUs.
We use the Adam optimizer and adopt batch sizes of 3, 5, and 3 for the OPV2V,  DAIR-V2X, and V2XSet datasets, respectively, with 60 epochs for each.
All models use the PointPillars~\cite{lang2019pointpillars} backbone to extract features from raw point clouds.
The confidence ratio $p$ selected for the CFG module is 70\% for OPV2V, 60\% for DAIR-V2X, and 70\% for V2XSet, and the surrounding grid size chosen for the MOF is 3×3.
To closely simulate real traffic conditions, we introduce localization and heading errors with standard deviations of 0.2 $m$ and 0.2°, respectively, sampled from a Gaussian distribution.
The communication range is limited to 70$m$, excluding agents beyond this range from collaborative communication.

\begin{table}
    \centering
    \caption{\footnotesize Collaborative perception performance on the OPV2V, DAIR-V2X and V2XSet dataset with a time delay of 100 $ms$, localization errors of 0.2 $m$, and heading errors of 0.2° using Average Percision(AP)@0.7/0.5 as metrics. \textbf{Bold numbers} indicate the best results, while \underline{underlined numbers} represent the second-best results. * represents a variation of PointPillar~\cite{lang2019pointpillars}.}
    \vspace{-10pt}
    \small
    \scalebox{0.85}{
    \begin{tabularx}{\textwidth}{>
    {\centering\arraybackslash}c|>
    {\centering\arraybackslash}c|>{\centering\arraybackslash}c|>{\centering\arraybackslash}c}
         \cline{1-4}
         \multirow{2}{*}{\textbf{Models}} & \textbf{OPV2V} &\textbf{DAIR-V2X} & \textbf{V2XSet} \\
         \cline{2-4}
        &  AP@0.7/0.5  &   AP@0.7/0.5  &  AP@0.7/0.5  \\
         \cline{1-4} 
         No Fusion*~\cite{lang2019pointpillars}&   48.66/68.71 & 43.57/50.03 &40.20/60.60\\
         \cline{1-4}
         Late Fusion*~\cite{lang2019pointpillars}& 59.48/79.62 &  34.47/51.14  & 30.75/54.92\\
         \cline{1-4}
         Intermediate Fusion*~\cite{lang2019pointpillars}& \underline{70.82}/\underline{ 88.41}  &  39.38/56.22  &59.38/83.18\\
         \cline{1-4}   
         When2com~\cite{liu2020when2com}&  57.55/74.11&  33.68/48.20 &41.85/67.41\\
         \cline{1-4}
         DiscoNet~\cite{li2021learning} & 68.64/84.72  &  40.69/52.67 &54.11/79.82 \\
         \cline{1-4}   
         Where2comm~\cite{hu2022where2comm} &69.73/85.16& 43.71/59.52&\underline{63.77}/83.17 \\
         \cline{1-4}
         V2X-ViT~\cite{xu2022v2x}& 70.06/84.65 &    40.43/53.08  & 61.49/\underline{83.63}\\
         \cline{1-4}
         Select2col~\cite{liu2024select2col}& 62.46/82.30 &   34.82/51.96  & 51.22/76.88\\
         \cline{1-4}
          ERMVP~\cite{zhang2024ermvp}& 69.71/86.63  &  \underline{46.96}/\underline{64.21}& 58.44/81.54 \\
         \cline{1-4}
        \textbf{ Ours} & \textbf{78.11}/\textbf{ 88.93}  & \textbf{48.27}/\textbf{ 65.12} & \textbf{65.86}/\textbf{84.40} \\
         \cline{1-4}
    \end{tabularx}
    }
    \label{tab:perception performance}
    \vspace{-15pt}
\end{table}

\begin{figure*}[!h]
    \begin{center}
   \includegraphics[width=0.95\linewidth]{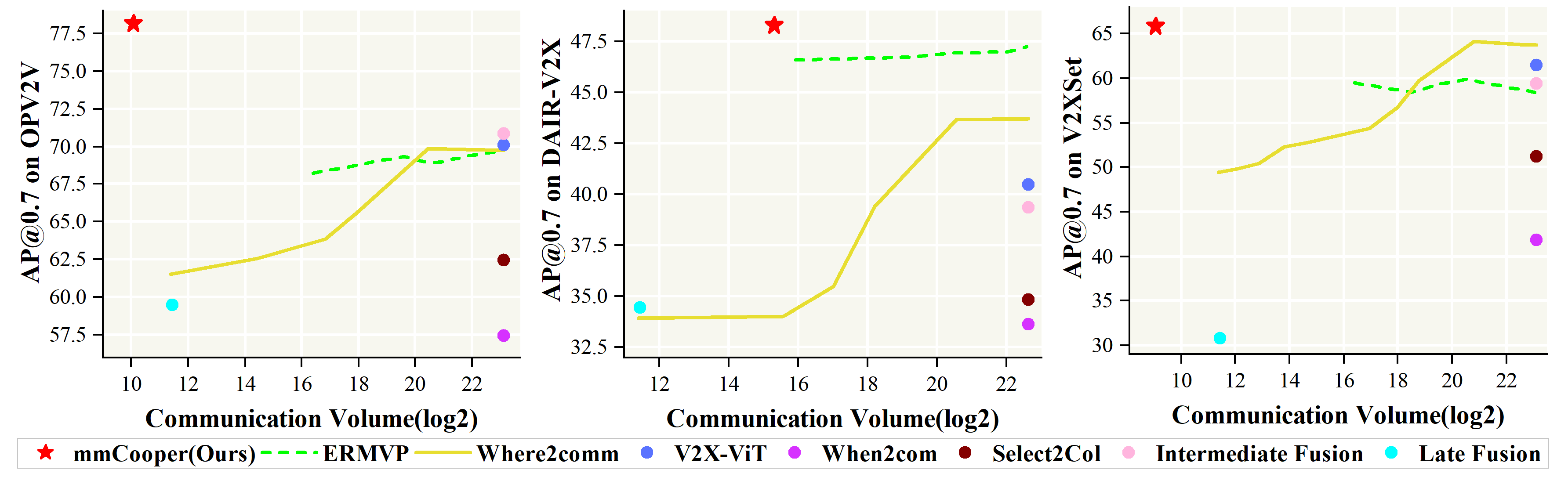}
    \vspace{-10pt}
    \caption{\footnotesize 
    Collaborative perception performance and communication volumes of all the models on the OPV2V, DAIR-V2X, and V2XSet datasets.}
    \label{fig:bandwidth-accuracy}    
    \end{center}
    \vspace{-25pt}
\end{figure*}

\subsection{Quantitative Evaluation}
\vspace{-2pt}
\textbf{Detection Performance.}
\cref{tab:perception performance} presents the comparison of the perception performance of our proposed mmCooper model and various baseline models on OPV2V,  DAIR-V2X, and V2XSet datasets with a time delay of 100 $ms$, localization errors of 0.2 $m$, and heading errors of 0.2°.
The No Fusion relies solely on ego agent observations.
Late Fusion shares only the bounding box results among agents.
Intermediate Fusion allows agents to share complete features extracted from raw point clouds.
The above three models are all variants based on the PointPillar~\cite{lang2019pointpillars} point cloud detector.
Additionally, we consider SOTA models, including When2com~\cite{liu2020when2com}, 
DisoNet~\cite{li2021learning},
Where2comm~\cite{hu2022where2comm}, V2X-ViT~\cite{xu2022v2x}, Select2col~\cite{liu2024select2col} and ERMVP~\cite{zhang2024ermvp}.
As shown in the table, our proposed mmCooper outperforms all the baselines on both simulated and real-world datasets.
Compared to existing SOTA models, mmCooper outperforms the second-best baseline on the OPV2V, DAIR-V2X, and V2XSet datasets by 7.29\%/0.59\%, 1.31\%/0.91\% and 2.09\%/0.77\% in AP@0.7/0.5, respectively.
This demonstrates the effectiveness and superiority of our multi-stage fusion framework.

\noindent \textbf{Communication Volume. }
\cref{fig:bandwidth-accuracy} illustrates the model performance under different bandwidth conditions on the OPV2V, DAIR-V2X, and V2XSet datasets.
Experimental results show that mmCooper significantly reduces communication costs, achieving performance levels comparable to late fusion.
In comparison to other SOTA models, the communication volume required to achieve optimal performance is 9153/156/18305 times more than our method.
Notably, on the OPV2V and V2XSet datasets, our method even achieves lower communication volume than late fusion.
This is because our CFG module suppresses low-confidence information from the transmission and selectively transmits complementary and exclusive information at intermediate and late stages rather than transmitting through only one stage encompassing all information of the entire scenario.
In addition, since some Late Fusion methods (e.g., OpenCOOD~\cite{xu2022opv2v}) transmit non-processed (e.g., NMS) BBoxes, our mmCooper even achieves a lower communication volume compared to the late fusion model on certain datasets.

\begin{table*}
    \centering
    \caption{\footnotesize Results for different localization errors ($m$) and transmission delay ($ms$) on the OPV2V and DAIR-V2X dataset evaluated using AP@0.7. \textbf{Bold numbers} indicate the best results, while \underline{underlined numbers} represent the second-best results.}
    \vspace{-7pt}
    \scalebox{0.9}{
    \begin{tabular}{c|ccc|ccc|ccc|ccc}
         \hline
          Noise Type& \multicolumn{6}{c|}{Localization Errors(m)} & \multicolumn{6}{c}{Transmission Delay(ms)}\\
            \hline
           Datasets & \multicolumn{3}{c|}{OPV2V} &\multicolumn{3}{c|}{DAIR-V2X} & \multicolumn{3}{c|}{OPV2V} &\multicolumn{3}{c}{DAIR-V2X}\\
         \hline
        Noise Level &  0.0&  0.2   & 0.4&0.0&0.2&0.4  &  0&  200   & 400&0&200&400\\
         \hline
         \hline
         No Fusion& 48.66 & 48.66 & 48.66 &43.57&43.57&43.57 &  48.66& 48.66 & 48.66 &43.57&43.57&43.57 \\
         When2com~\cite{liu2020when2com}&70.82  & 64.35 & 59.98&39.50 & 36.62 & 35.03  & 70.82 & 43.96 & 36.55&39.50&36.83&33.72 \\
         DiscoNet~\cite{li2021learning}& 76.05 & 73.59 & 65.85 & 46.94& 46.03&44.96  & 76.05 & 60.08 & 50.33 &46.94&44.03&41.57 \\
         Where2comm~\cite{hu2022where2comm}& 78.47 & 75.45 & 69.77&52.34&\underline{49.34}&\underline{46.96}  & 78.47 & 65.23 & 53.33&52.34&47.76&45.25\\
         V2X-ViT~\cite{xu2022v2x}& 77.83 & 75.21 & 66.72&46.12&43.81&42.53 & 77.83 & 66.08 & 50.33&46.12&45.49&43.99\\
         ERMVP~\cite{zhang2024ermvp}& \underline{80.55} & \underline{76.19} &  \underline{72.78} &\underline{53.27}&48.66&45.97 &\underline{80.55}  & \underline{68.89}&\underline{68.31}&\underline{53.27}&\underline{49.68}&\underline{48.60}\\
         \textbf{Ours}& \textbf{86.41} & \textbf{82.53} & \textbf{76.80}&\textbf{56.06}&\textbf{51.52}& \textbf{47.66} & \textbf{86.41} & \textbf{77.44} & \textbf{75.26}&\textbf{56.06}&\textbf{50.96}& \textbf{50.81}\\
         \hline
    \end{tabular}
    }
    \label{tab:robustness}
    \vspace{-15pt}
\end{table*}

\noindent \textbf{Robustness to Localization Errors and Transmission Delay.}
We verified the robustness of the agents during collaborative processes in the presence of localization errors and transmission delay on the OPV2V and DAIR-V2X datasets. 
Following the noise settings in~\cite{xu2022v2x}, localization noise was sampled from a Gaussian distribution with a mean of 0 $m$ and a standard deviation $\sigma \in \{0, 0.2, 0.4\} m $.
As shown in \cref{tab:robustness}, as the standard deviation of the localization noise gradually increases, the detection performance of the models decreases.
On the DAIR-V2X dataset, when2com even performs worse than the No Fusion method when the localization error exceeds 0.2 $m$.
Our method outperforms all the baselines under all localization noise settings, demonstrating its robustness to varying levels of localization error.
Transmission delay can lead to misalignment of features and proposals, thereby impacting detection performance.
We evaluated the models on the OPV2V and DAIR-V2X datasets with time delays $\tau \in \{0,200,400\}ms$ set to 0 $ms$, 200 $ms$, and 400 $ms$. 
With increasing time delay, the detection performance of all methods gradually degrades.
However, mmCooper achieves the highest Average Precision (AP) over all the models on both datasets.
This robustness to localization errors and transmission delay demonstrates the effectiveness of the MOF and BFC modules in addressing the potential misalignment and errors during communication among agents.
For more results regarding errors during transmission, please refer to the supplementary.

\subsection{Ablation Study}
\vspace{-2pt}

\begin{table}

\caption{Ablation study results of different designs in mmCooper on the OPV2V and DAIR-V2X datasets. CFG: Confidence-based Filter Generation Module; MOF: Multi-scale Offset-aware Fusion; BFC: BBox Filtering \& Calibration Module; LF: Late-stage Fusion; IF: Intermediate-stage Fusion.  }
\vspace{-7pt}
\centering
\resizebox{0.48\textwidth}{!}{
\renewcommand{\arraystretch}{1.2}

    \begin{tabular}{c c c |c c | c |c}
     \hline
     \multirow{2}{*}{\textbf{CFG}} & \multirow{2}{*}{\textbf{MOF}} &\multirow{2}{*}{\textbf{BFC}}&\multirow{2}{*}{\textbf{ LF}}&\multirow{2}{*}{\textbf{ IF}}& \multicolumn{2}{c}{\textbf{AP@0.7/0.5$(\uparrow)$}} \\
     \cline{6-7}
     &&&& &  OPV2V  & DAIR-V2X\\
     \hline
     \textcolor{green}{\ding{52}}  &\textcolor{green}{\ding{52}} & \textcolor{green}{\ding{52}}&\textcolor{green}{\ding{52}}& \textcolor{green}{\ding{52}} & \textbf{78.11/88.93}& \textbf{48.27/65.12}  \\
     \hline
     \multicolumn{7}{c}{Importance of Core Components}  \\
     \hline     
     \textcolor{red}{\ding{56}}&\textcolor{green}{\ding{52}}&\textcolor{green}{\ding{52}}&\textcolor{gray}{\textbf{--}}&  \textcolor{gray}{\textbf{--}}&73.46/88.84& 47.66/63.63 \\
     \hline
   \textcolor{green}{\ding{52}}  &\textcolor{red}{\ding{56}}&\textcolor{green}{\ding{52}}  &\textcolor{gray}{\textbf{--}}  &\textcolor{gray}{\textbf{--}}  & 72.60/88.51& 47.27/63.49 \\
     \hline
   \textcolor{green}{\ding{52}}  &\textcolor{green}{\ding{52}}& \textcolor{red}{\ding{56}} &\textcolor{gray}{\textbf{--}} & \textcolor{gray}{\textbf{--}} & 76.77/88.56 & 44.31/59.68\\
     \hline
     \multicolumn{7}{c}{Results of Single-stage Fusion}  \\
     \hline
       \textcolor{gray}{\textbf{--}} & \textcolor{gray}{\textbf{--}} & \textcolor{gray}{\textbf{--}} &\textcolor{red}{\ding{56}}&\textcolor{green}{\ding{52}}&76.83/88.65& 46.65/62.84\\
     \hline
     \textcolor{gray}{\textbf{--}} & \textcolor{gray}{\textbf{--}} & \textcolor{gray}{\textbf{--}} &\textcolor{green}{\ding{52}} & \textcolor{red}{\ding{56}} &66.80/78.62& 45.78/56.88\\
     \hline
    \end{tabular}
    }
    \label{tab:Ablation study}
    \vspace{-10pt}
\end{table}

We conducted comprehensive ablation studies on the OPV2V and DAIR-V2X datasets to demonstrate the importance of different components, as shown in \cref{tab:Ablation study}.

\noindent \textbf{Importance of Core Components.}
All the core components contribute to performance improvement.
Removing the CFG module results in the transmission of information for both the intermediate and late stages across all locations without suppression or selection.
While this variant increases the amount of transmitted information compared to mmCooper, it leads to a performance drop.
This highlights the importance of filtering out low-confidence information and selectively transmitting data at different stages to optimize performance.
The absence of the MOF module replaces the MOA module with a self-attention mechanism, while removing the BFC module eliminates the filtering and refinement of received bounding boxes. 
Both modifications result in a performance drop, highlighting the importance of incorporating spatial neighborhood information to mitigate misalignment and applying bounding box filtering and refinement to address redundancy, imprecision, or potential errors in received detections.



\noindent \textbf{Superiority of Multi-stage Fusion.}
We entirely removed the Late Fusion (LF) or the Intermediate Fusion (IF) to degrade our model to single-stage fusion.
The experimental results indicate that removing either module leads to a performance decline, underscoring the effectiveness of the proposed multi-stage cooperation framework.

\begin{figure}[!t]
    \begin{center}

    \includegraphics[width=1\linewidth]{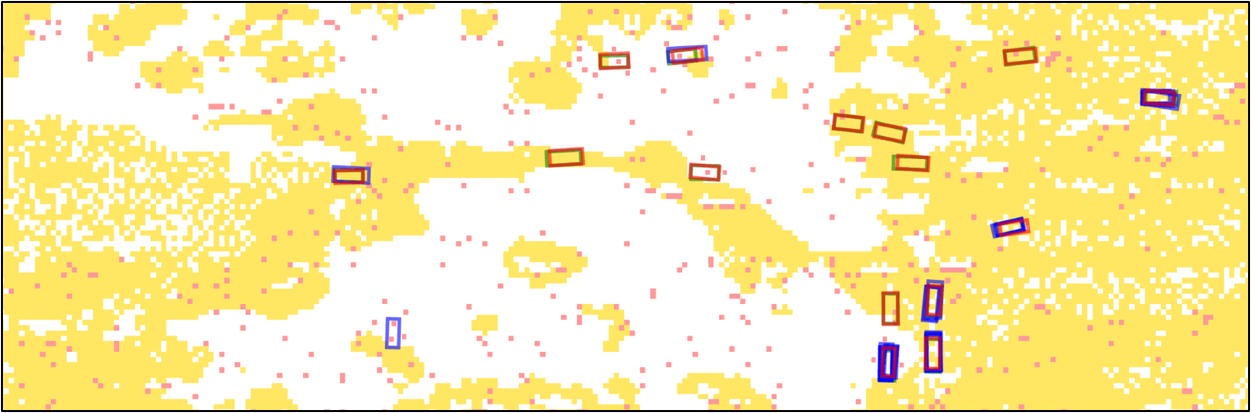}
    \vspace{-20pt}
    \caption{\footnotesize Visualization of two-stage fusion on DAIR-V2X. We show the results from the Confidence-based Filtering Generation Module, including discarded information (white background), BBoxes for transmission (red dots), and features for transmission (yellow background). We also show the BFC module results, including uncalibrated BBoxes (blue boxes), calibrated BBoxes (red boxes), and ground truth BBoxes (green boxes).}
    \label{fig:det_vis}    
    \vspace{-28pt}
    \end{center}
\end{figure}

\subsection{Qualitative Evaluation}

\noindent \textbf{Visualization of Two-stage Fusion.}
As shown in the \cref{fig:det_vis}, we visualize the transmission decisions and bounding boxes refinement in the two-stage fusion on the DAIR-V2X dataset.
The yellow background and red dots show the selective transmission of data through two stages.
The white background shows the locations discarded by the CFG module, showing a substantial amount of low-confidence or irrelevant information in the scenario.
The removal and correction of the uncalibrated blue boxes demonstrates the BFC module’s effectiveness in filtering and refining received bounding boxes by eliminating redundancies and improving alignment with the ground truth.

\begin{figure}
    \begin{center}
    \includegraphics[width=0.9\linewidth]{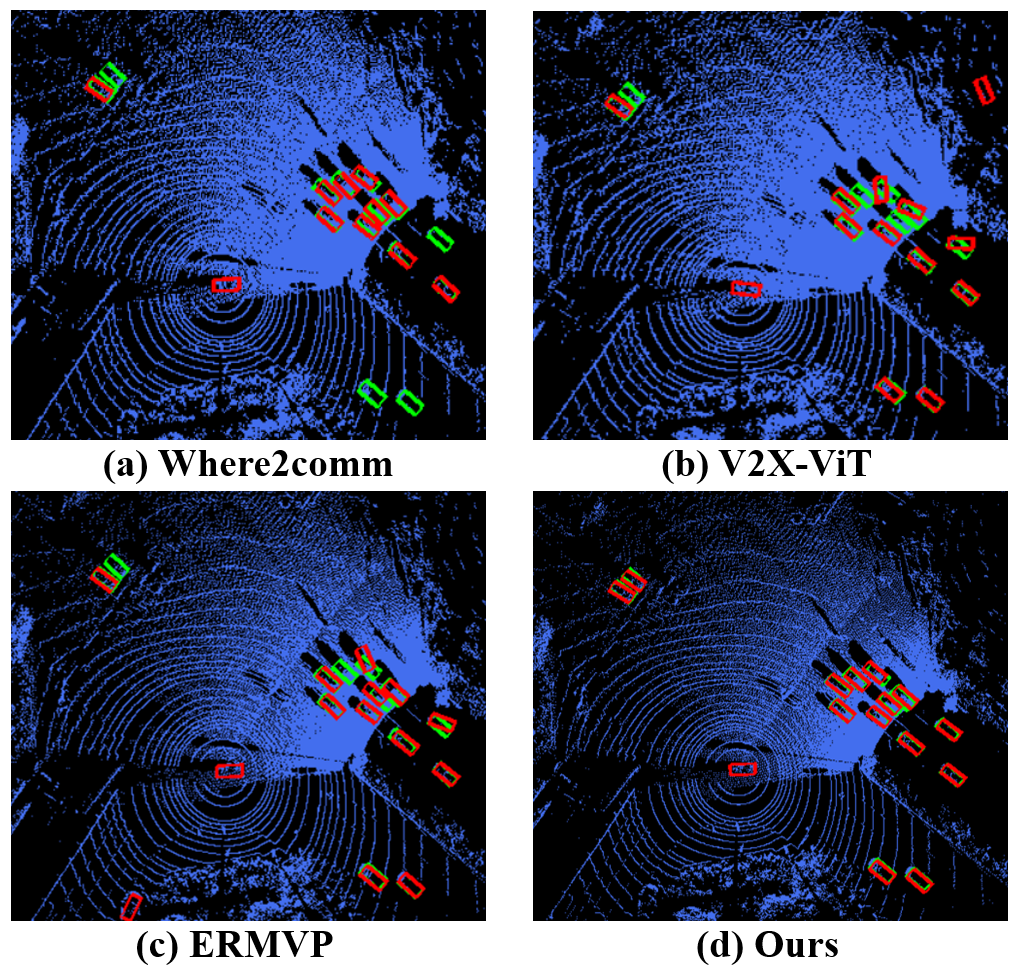}
    \vspace{-10pt}
    \caption{\footnotesize Visualization comparison of detection results on the DAIR-V2X dataset. Green and red boxes represent the ground truth and the model-predicted bounding boxes, respectively.}
    \label{fig:vis}     
    \vspace{-17pt}
    \end{center}
\end{figure}

\noindent \textbf{Visualization of Detection Results.}
\cref{fig:vis} shows visualization results for a scenario in the DAIR-V2X dataset.
Compared to baseline methods, mmCooper demonstrates high-precision 3D object detection, accurately predicting nearly all ground truth objects. 
In contrast, baseline methods either fail to detect certain vehicles or produce inaccurate bounding boxes. †Refer to the supplementary materials for additional visualized detection results.


%% file: sec/5_conclusion.tex
\vspace{-3pt}
\section{Conclusion}
\vspace{-3pt}
In this paper, we have proposed mmCooper, a novel multi-agent, multi-stage, communication-efficient, and collaboration-robust cooperative perception framework.
mmCooper is the first framework to achieve adaptive fusion of complementary data across stages for cooperative perception.
We introduce the Confidence-based Filter Generation Module to enable dynamic partitioning of intermediate features and late-stage bounding boxes, balancing communication load and perception performance. Additionally, the Multi-scale Offset-aware Fusion module and BBox Filtering \& Calibration module are incorporated to address potential misalignment and calibration noise among agents. 
mmCooper outperforms the second-best state-of-the-art models on the OPV2V, DAIR-V2X, and V2Xset datasets, achieving improvements of 7.29\%, 1.31\%, and 2.09\% in AP@0.7, respectively, while reducing communication volume by factors of 9153, 156 and 18305.

\noindent
\textbf{Acknowledgements.} This work is supported by the National Natural Science Foundation of China under Grant 62272357 and 62302326, Wuhan Science and Technology Project for Key Research and Development under Grant 2024050702030090 and Wuhan Science and Technology Joint Project for Building a Strong Transportation Country under Grant 2024-2-7.

%% file: sec/supp.tex
\clearpage
\setcounter{page}{1}
\maketitlesupplementary
\setcounter{section}{0}

\section{Overview}
The supplementary material is organized into the following sections:
\begin{itemize}
\item \cref{sec:pipeline}: The System Pipeline of mmCooper
\item \cref{sec:multi fusion details}: Details of Multi-stage Fusion Method
\item \cref{sec:two_datasets}: Additional Experimental Results
  \begin{itemize}
      \item \cref{sec:implementation}: Implementation Details 
      \item \cref{sec:supp_localization}: Supplements on Localization Errors
      \item \cref{sec:supp_delay}: Supplements on Transmission Delays
      \item \cref{sec:heading}: Robustness to Heading Errors
      \item \cref{sec:ablation on v2xset}: More Ablation on V2VSet Dataset
      \item \cref{sec:v2v4real result}: Performance on V2V4Real Dataset
      \item \cref{sec:computation costs}: Computation Costs
      \item \cref{sec:DBA}: Ablation of Deformable BBox Attention (DBA)
      \item \cref{sec:number of agents}: Impact of Varying the Number of Agents
  \end{itemize}
\item \cref{sec:add_qualitative}: Additional Qualitative Evaluation Results
  \begin{itemize}
      \item \cref{subsec:det}: Visualization of Detection Result 
      \item \cref{subsec:multi fusion}: Visualization of Multi-stage Fusion
  \end{itemize}
  
\end{itemize}
\section{The System Pipeline of mmCooper}
\label{sec:pipeline}
The proposed system pipeline of mmCooper is illustrated in \cref{Alg:alg1}. Note that the following pipeline is executed in parallel across all agents. For simplicity, we describe the pipeline from the perspective of the ego agent. The ego agent is represented by $i$, while the collaborative agents are denoted by $j$.
First, the agent generates BEV features \( F_{}^{i} \) and \( F_{}^{j} \) through the Observation Encoding. In the Information Broadcasting, agents generate initial coarse bounding boxes \( B_{}^{i} \) and \( B_{}^{j} \), where \( N_{b}^{i} \) and \( N_{b}^{j} \) represent the number of bounding boxes predicted by the ego agent and the collaborating agents, respectively. The collaborative agents then package and broadcast the filtered BEV features along with the coarse bounding boxes. Specifically, \( f_{gaussian}(\cdot) \) denotes a Gaussian filter, and \( \{ \hat{F}_{}^{j}, \hat{\mathcal{B}}_{}^{j} \} \) represents the filtered features and bounding boxes from the collaborative agents.

Subsequently, in the Intermediate-stage Fusion, the ego agent performs feature fusion using the Multi-scale Offset-aware Attention module. The outputs from the fused features at different scales, after undergoing upsampling, are concatenated to obtain the final fused features. Here, \( f_{up2}(\cdot) \) and \( f_{up3}(\cdot) \) denote the upsampling operations, while \( \{ F_{sc1}^{i}, F_{sc2}^{i}, F_{sc3}^{i} \} \) represent the feature fusion outputs at three different scales.

In the later-stage fusion, the BBox Filtering \& Calibration Module is employed to learn the bounding box offsets and scores. Specifically, DBA refers to the Deformable Bounding Boxes Attention Module, while FFN denotes the feed-forward network, \( f_{enc,b}(\cdot) \) represents the feature extractor for bounding boxes, \( f_{off}(\cdot) \) is the offset mapping function, and \( f_{score}(\cdot) \) is the score mapping function. $\varphi(\cdot)$ represents the filtering and calibration through scores and offsets.

Finally, the fused features are input into a detection head to predict the fused bounding box \( \mathcal{B}_{fused} \). The final bounding box is composed of the bounding box predicted from the fused features  \( \mathcal{B}_{fused} \) and the bounding box obtained after filtering and calibration  \( \hat{\mathcal{B}}_{m}^{j} \)from collaborative agents. \( f_{post}(\cdot) \) represents the combination of all bounding boxes.

\begin{algorithm*}

\caption{System Pipeline of the Proposed mmCooper}
Define $ N = \{1,...,n\}$ as the agent set, $i \in N$ represents the ego agent, while $j\in N$ denotes the collaborative agents. $x_{i/j}$ serves as the input point cloud. 
\label{alg:111}

---

\textcolor{blue}{\textbf{\# For collaborative agents.}}

\textcolor{blue}{\# Observation Encoding.} 

\textbf{for} each agent $ j \in N$, \textbf{do}

\hspace*{1.0em}  $F_{}^{j} = \psi_{encoder}(x_{}^{j}) \in \mathbb{R}^{C\times 
H\times W}$

\textbf{end for }



\textcolor{blue}{\# Information Broadcasting.}

\textbf{for} each $j \in N$, \textbf{do}

\hspace*{1.0em} $\mathcal{B}_{}^{j} = f_{dec}(F_{}^{j}) \in \mathbb{R}^{ N_{b}^{j}\times 7}$

\hspace*{1.0em} $C_{f}^{j}, C_{b}^{j} = \phi _{conf} \left (F_{}^{j} \right )$

\hspace*{1.0em} $C_{f}^{j}, C_{b}^{j} = f_{gaussian} \left (C_{f}^{j}, C_{b}^{j} \right )$

\hspace*{1.0em} $M_{f,t}^{j}, M_{b,t}^{j} = f_{top} \left (C_{f}^{j}, C_{b}^{j} \right )$

\hspace*{1.0em} $M_{f,g}^{j}, M_{b,g}^{j} = f_{max}  \left(\frac{exp\left (( logC_{f/b}^{j} + g_{f} )/\tau\right ) }{ { {\textstyle \sum_{s}} exp\left (( logC_{s}^{j} + g_{s} )/\tau\right )}  } \right) $

\hspace*{1.0em} $\mathcal{M}_{f/b}^{j} = M_{f/b,g}^{j} \odot M_{f/b,t}^{j}$ 

\hspace*{1.0em} $\hat{F}_{}^{j}=\mathcal{M}_{f}^{j} \odot F_{}^{j} $;  $\hat{\mathcal{B}}_{}^{j}=\mathcal{M}_{b}^{j} \odot \mathcal{B}_{}^{j}$

\hspace*{1.0em} broadcast $\{\hat{F}_{}^{j}, \hat{B}_{}^{j}\}$ to other agent

\textbf{end for}

---

\textcolor{blue}{\textbf{\# For ego agent.}}

\textcolor{blue}{\# Observation Encoding.} 

$F_{}^{i} = \psi_{encoder}(x_{}^{i}) \in \mathbb{R}^{C\times H\times W}$

\textcolor{blue}{\# Information Broadcasting.}

$\mathcal{B}_{}^{i} = f_{dec}(F_{}^{i}) \in \mathbb{R}^{ N_{b}^{i}\times 7}$

 \textcolor{blue}{\# Intermediate-stage Fusion.}

Receive $\{\hat{F}_{}^{j}, \hat{\mathcal{B}}_{}^{j}\}$ sent by collaborative agents.

Encode the feature set $\{F_{}^{i}, \hat{F}_{}^{j}\}$ into three scales $S = \{sc1,sc2,sc3\}.$

\textbf{For} each $sc \in S$, \textbf{do}

\hspace*{1.0em}  $\mathcal{F}_{sc,(r,c)}^{i} =  CrossAttn(MLP( F_{sc,(r,c)}^{i}), \hat{F}_{sc,(r,c)}^{j, s\times s}, \hat{F}_{sc,(r,c)}^{j, s\times s})$   

\textbf{end for}

$\mathcal{F}_{}^{i}=concat(\mathcal{F}_{sc1}^{i}, f_{up2}(\mathcal{F}_{sc2}^{i}), f_{up3}(\mathcal{F}_{sc3}^{i}) )$

 \textcolor{blue}{\# Late-stage Fusion.}

$F_{b}(q) = f_{enc,b}(\hat{\mathcal{B}}_{}^{j})$

$DBA(q)=\sum_{\alpha =1}^{A} W_{\alpha }[\sum_{n=1}^{N} \sum_{m=1}^{M} \omega (W_{\beta }F_{b}(q))\mathcal{F}_{}^{i}(q+\bigtriangleup q_{m})] + F_{b}(q)$

$F_{DBA} = FFN(F_{DBA}) + F_{DBA}$

$off,score = f_{off}(F_{DBA}), f_{score}(F_{DBA})$

$\hat{\mathcal{B}}_{m}^{j} = \varphi (off, score, \hat{\mathcal{B}}_{}^{j})$

 \textcolor{blue}{\# Detection Decoders}

$\mathcal{B}_{fused} = f_{dec}(\mathcal{F}_{}^{i})$

$\mathcal{B}_{final}^{i} = f_{post}(\mathcal{B}_{}^{i} , \hat{\mathcal{B}}_{m}^{j} , \mathcal{B}_{fused})$
\label{Alg:alg1}
\end{algorithm*}

\section{Details of Multi-stage Fusion Method}
\label{sec:multi fusion details}
\begin{figure}[h]
    \small
    \begin{center}
    \includegraphics[width=1.0\linewidth]{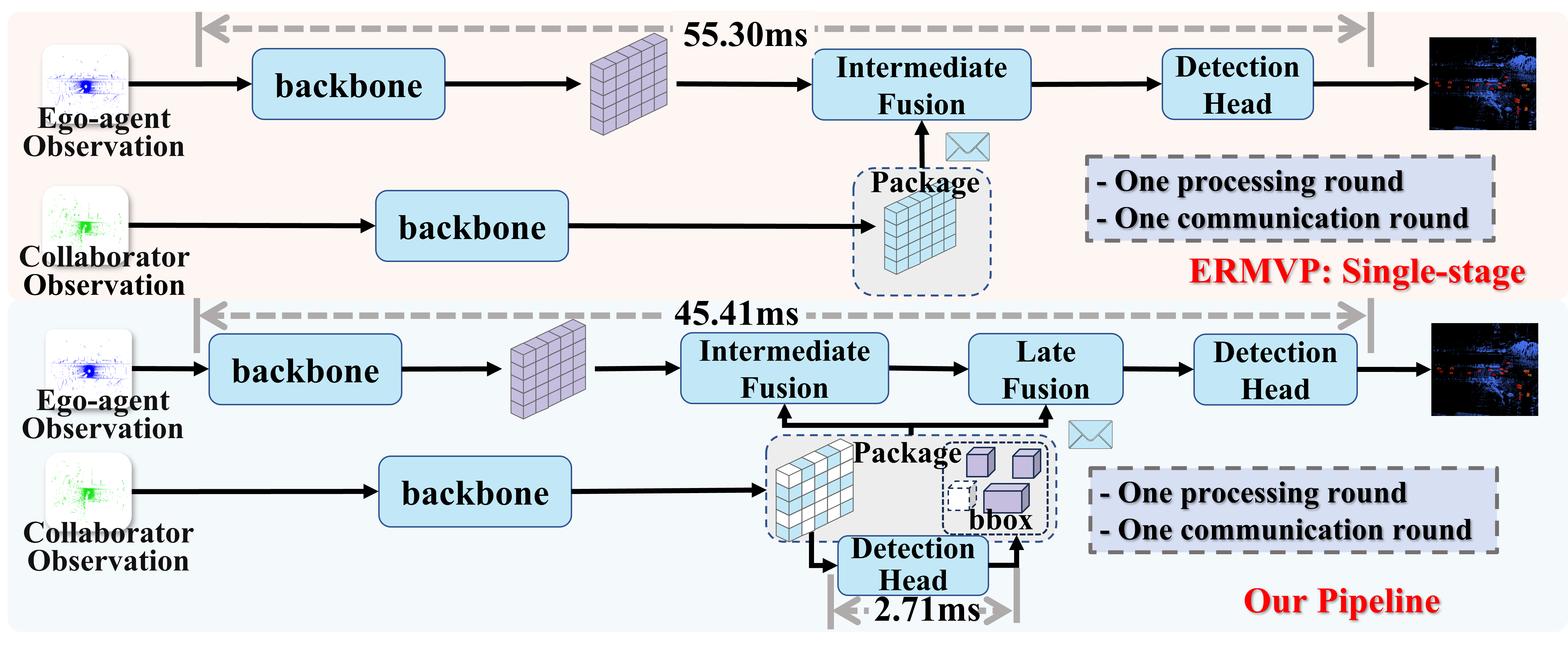}
    \caption{Comparison of single-stage method and our method.}
    \vspace{-12pt}
    \label{fig:stage}   
    \end{center}
\end{figure}

Note that our proposed model requires only a single round of processing and communication, the same as other single-stage methods.
As shown in \cref{fig:stage}, both intermediate-stage features and late-stage bboxes are handled within a single inference cycle and jointly transmitted in one communication round.
While mmCooper includes two detection heads, they share parameters, and each head introduces only 2.71ms of processing time.
Notably, mmCooper is more efficient than ERMVP, a single-stage intermediate fusion method, with a lower total runtime of 45.41ms compared to ERMVP’s 55.30ms.
Combined with its superior performance, this confirms that mmCooper remains both practically feasible and system-efficient, without the overhead suggested in the comment.

\section{Additional Experimental Results}
\label{sec:two_datasets}

In this section, we provide additional experiments to supplement the results on the datasets OPV2V~\cite{xu2022opv2v}, DAIR-V2X~\cite{yu2022dair}, V2XSet~\cite{xu2022v2x} and V2V4Real~\cite{xu2023v2v4real} presented in the \textit{main paper}.

\subsection{Implementation Details}
\label{sec:implementation}
On the OPV2V, DAIR-V2X and V2XSet datasets, the dimensions of the voxels encoded by the encoder are $0.4 \times 0.4 \times 4$. The shape of the shared BEV features among agents is (64, 100, 252) for DAIR-V2X and (64, 100, 352) for OPV2V and V2XSet. The shared bounding boxes among agents are represented by their center coordinates, dimensions (length, width, height), and heading angle. The detection decoder consists of two distinct 1×1 convolutional layers.

\subsection{Supplements on Localization Errors}
\label{sec:supp_localization}
\begin{figure*}
    \begin{center}
    \includegraphics[width=\linewidth]{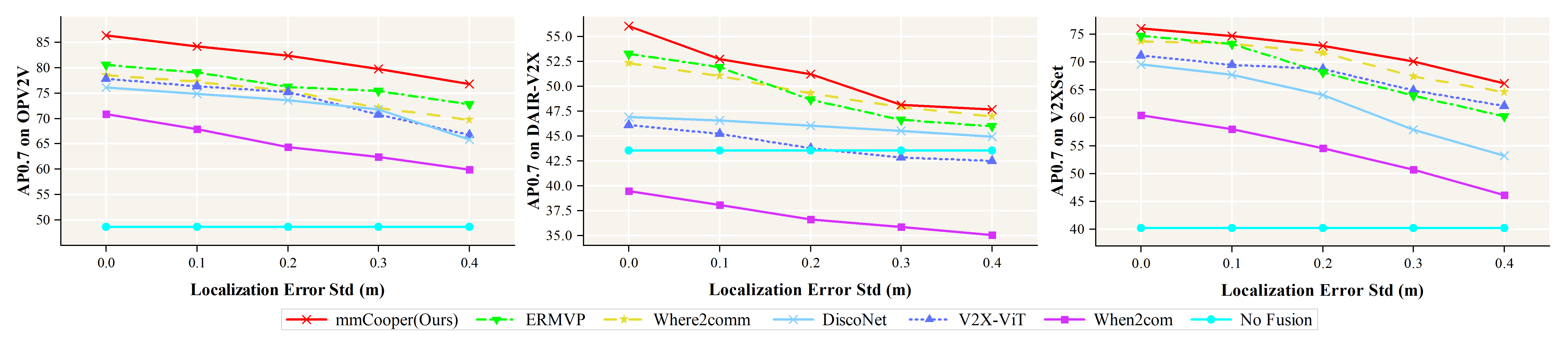}
    \vspace{-10pt}
    \caption{Robustness to the localization error on the OPV2V, DAIR-V2X and V2XSet datasets.}
    \label{fig:localization}     
    \vspace{-18pt}
    \end{center}
\end{figure*}

The localization errors on the OPV2V, DAIR-V2X and V2XSet datasets are sampled from a Gaussian distribution with a mean of 0 $m$ and a standard deviation $\sigma\in\{0.0, 0.1, 0.2, 0.3, 0.4\} m $. The experimental results, as shown in \cref{fig:localization}, demonstrate that our proposed mmCooper outperforms the existing state-of-the-art methods~\cite{liu2020when2com,hu2022where2comm,li2021learning,xu2022v2x,zhang2024ermvp} across different levels of localization error, highlighting its robustness to such errors.

\subsection{Supplements on Transmission Delays}
\label{sec:supp_delay}
\begin{figure*}
    \begin{center}
    \includegraphics[width=\linewidth]{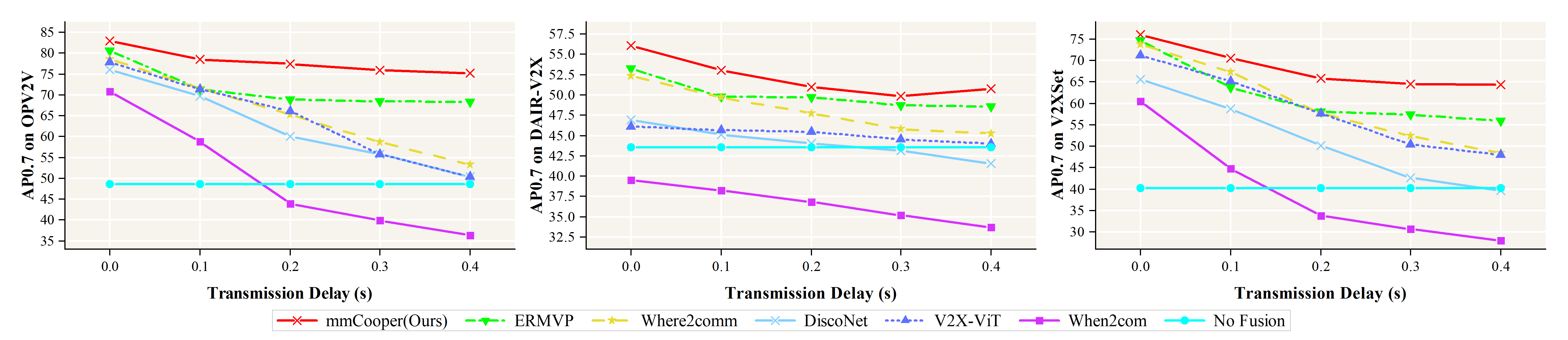}
    \vspace{-10pt}
    \caption{Robustness to the transmission delay on the OPV2V, DAIR-V2X and V2XSet datasets.}
    \label{fig:transmission}     
    \vspace{-18pt}
    \end{center}
\end{figure*}

We evaluated the performance of the models on the OPV2V, DAIR-V2X and V2XSet datasets under transmission delays for $\{0, 100, 200, 300, 400\} ms$. As shown in \cref{fig:transmission}, our proposed mmCooper consistently outperforms the existing state-of-the-art methods across all transmission delay conditions, demonstrating the superiority of our approach in scenarios with transmission delays.

\subsection{Robustness to Heading Errors.}
\label{sec:heading}
\begin{figure*}
    \begin{center}
    \includegraphics[width=\linewidth]{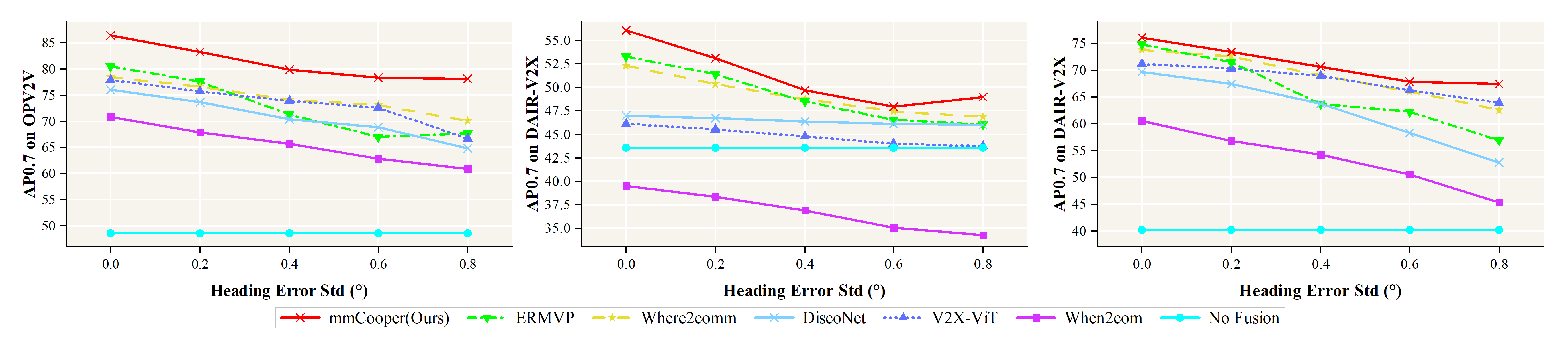}
    \vspace{-10pt}
    \caption{Robustness to the heading error on the OPV2V, DAIR-V2X and V2XSet datasets.}
    \label{fig:heading}     
    \vspace{-18pt}
    \end{center}
\end{figure*}

\cref{fig:heading} demonstrates the performance of our proposed mmCooper method compared to other baseline methods under varying levels (i.e., $\{0.0, 0.2, 0.4, 0.6, 0.8\}^{\circ}$) of heading error noise on the OPV2V, DAIR-V2X and V2XSet datasets. As illustrated in \cref{fig:heading}, the performance of all models decreases as the heading errors increase. However, mmCooper consistently outperforms the current state-of-the-art models across all error levels, highlighting the advantages of our designed Multi-Scale Offset-Aware Fusion Module and BBox Filtering \& Calibration Module in enhancing system robustness.

\subsection{More Ablation on V2XSet Dataset}
\label{sec:ablation on v2xset}
We supplement the \textit{main paper} with ablation experiments conducted on the V2XSet dataset. As shown in \cref{tab:v2xset ablation}, the results align with those presented in the \textit{main paper}, demonstrating that the absence of any key component leads to performance degradation. Moreover, downgrading mmCooper to a single-stage approach also results in decreased performance.

\begin{table}[h]
\caption{Ablation study results of different designs in mmCooper on the V2XSet datasets. CFG: Confidence-based Filter Generation Module; MOF: Multi-scale Offset-aware Fusion; BFC: BBox Filtering \& Calibration Module; LF: Late-stage Fusion; IF: Intermediate-stage Fusion.}
\vspace{-7pt}
\centering
\resizebox{0.48\textwidth}{!}{
\renewcommand{\arraystretch}{1.2}
    \begin{tabular}{c c c |c c | c }
     \hline
     \multirow{2}{*}{\textbf{CFG}} & \multirow{2}{*}{\textbf{MOF}} &\multirow{2}{*}{\textbf{BFC}}&\multirow{2}{*}{\textbf{ LF}}&\multirow{2}{*}{\textbf{ IF}}& \multicolumn{1}{c}{\textbf{AP@0.7/0.5$(\uparrow)$}} \\
     \cline{6-6}
     &&&& &  V2XSet \\
     \hline
     \textcolor{green}{\ding{52}}  &\textcolor{green}{\ding{52}} & \textcolor{green}{\ding{52}}&\textcolor{green}{\ding{52}}& \textcolor{green}{\ding{52}} & \textbf{65.86/84.40}  \\
     \hline
     \multicolumn{6}{c}{Importance of Core Components}  \\
     \hline     
     \textcolor{red}{\ding{56}}&\textcolor{green}{\ding{52}}&\textcolor{green}{\ding{52}}&\textcolor{gray}{\textbf{--}}&  \textcolor{gray}{\textbf{--}}&62.78/82.47 \\
     \hline
   \textcolor{green}{\ding{52}}  &\textcolor{red}{\ding{56}}&\textcolor{green}{\ding{52}}  &\textcolor{gray}{\textbf{--}}  &\textcolor{gray}{\textbf{--}}  & 64.33/84.04\\
     \hline
   \textcolor{green}{\ding{52}}  &\textcolor{green}{\ding{52}}& \textcolor{red}{\ding{56}} &\textcolor{gray}{\textbf{--}} & \textcolor{gray}{\textbf{--}} & 64.70/83.59 \\
     \hline
     \multicolumn{6}{c}{Results of Single-stage Fusion}  \\
     \hline
       \textcolor{gray}{\textbf{--}} & \textcolor{gray}{\textbf{--}} & \textcolor{gray}{\textbf{--}} &\textcolor{red}{\ding{56}}&\textcolor{green}{\ding{52}}&64.19/83.73\\
     \hline
     \textcolor{gray}{\textbf{--}} & \textcolor{gray}{\textbf{--}} & \textcolor{gray}{\textbf{--}} &\textcolor{green}{\ding{52}} & \textcolor{red}{\ding{56}} &54.36/72.14\\
     \hline
    \end{tabular}
    }
    \label{tab:v2xset ablation}
    \vspace{-10pt}
\end{table}

\begin{table}[h]
    \centering
    \caption{Collaborative perception performance on the V2V4Real dataset with a time delay of 100 $ms$, localization errors of 0.2 $m$, and heading errors of 0.2° using Average Percision(AP)@0.7/0.5 as metrics.}
    
    \scalebox{1.0}{
    \begin{tabular}{c|c}
         \cline{1-2}
         \multirow{2}{*}{\textbf{Models}} & \textbf{V2V4Real} \\
         \cline{2-2}
        &  AP@0.7/0.5   \\
         \cline{1-2} 
         No Fusion*~\cite{lang2019pointpillars}&   24.72/43.52\\
         \cline{1-2}
         Late Fusion*~\cite{lang2019pointpillars}& 17.79/39.91 \\
         \cline{1-2}
         Intermediate Fusion*~\cite{lang2019pointpillars}&28.77/49.19 \\
         \cline{1-2}   
         Where2comm~\cite{hu2022where2comm} &\underline{29.83}/\underline{49.43} \\
         \cline{1-2}
         V2X-ViT~\cite{xu2022v2x}& 25.08/47.85 \\
         \cline{1-2}
          ERMVP~\cite{zhang2024ermvp}& 26.81/44.57 \\
         \cline{1-2}
        \textbf{ Ours} & \textbf{31.47}/\textbf{50.21}  \\
         \cline{1-2}
    \end{tabular}
    }
    \centering
    \label{tab:v2v4rael_table}
\end{table}

\subsection{Performance on V2V4Real Dataset}
\label{sec:v2v4real result}

To further evaluate the performance of our model on more datasets, we also report the results on the V2V4Real dataset.
Our experimental setup on the V2V4Real dataset is consistent with that of the DAIR-V2X dataset. As shown in \cref{tab:v2v4rael_table}, mmCooper still demonstrates outstanding performance on the V2V4Real dataset.

\subsection{Computation Costs}
\label{sec:computation costs}
\cref{tab:inference time} presents the overall computation time of our model as well as the computation time of each module. Although both our intermediate-stage and late-stage fusion modules utilize attention-based mechanisms, due to the sparsity of intermediate-stage features from collaborators and the sparsity of reference points used in deformable attention during late-stage fusion, our model achieves inference time comparable to other models.

\subsection{Ablation of Deformable BBox Attention (DBA)}
\label{sec:DBA}
To further demonstrate the necessity of the Deformable BBox Attention (DBA) module, we replace it with the standard cross-attention mechanism. As shown in \cref{tab:DBA}, due to the lack of focus on key reference points, replacing DBA with standard cross-attention leads to performance drops of 4.07\%/0.07\%, 0.58\%/1.97\%, and 1.44\%/0.69\% in AP@0.7/0.5 on OPV2V, DAIR-V2X, and V2XSet respectively.

\begin{table*}[h]
    \centering 
    \hspace{0.02\textwidth}
    \begin{minipage}[t]{0.45\textwidth}
    \begin{tabular}{c|ccc|c}
         \cline{1-2}
         \multirow{2}{*}{\textbf{Models}} & \textbf{OPV2V}\\
         \cline{2-2}
         &  Inference Time(ms)\\
         \cline{1-2}
         Intermediate Fusion* & 10.07  \\
         \cline{1-2}   
         Where2comm~\cite{hu2022where2comm} &17.05 \\
         \cline{1-2}

         V2X-ViT~\cite{xu2022v2x}& 114.11 \\
         \cline{1-2}

         Select2col~\cite{liu2024select2col}& 27.00 \\
         \cline{1-2}

          ERMVP~\cite{zhang2024ermvp}& 55.30  \\
         \cline{1-2}

        \textbf{ Ours} & 45.41 \\
         \cline{1-2}
    \end{tabular}
    \end{minipage}
    \hspace{0.02\textwidth}
    \centering
    \begin{minipage}[t]{0.45\textwidth} 
    \renewcommand{\arraystretch}{1.14}
        \begin{tabular}{c|c}
            \hline
            \multirow{2}{*}{\textbf{Module}} & \textbf{OPV2V} \\
            \cline{2-2}
            &Inference Time(ms)\\
            \cline{1-2}
           Observation Encoding& 12.26 \\
            \cline{1-2}
            Information Broadcasting &14.37 \\
            \cline{1-2}
            Intermediate-stage Fusion& 13.54 \\
            \cline{1-2}
            Late-stage Fusion&4.06\\
            \cline{1-2}
            mmCooper&45.41\\
            \hline      
        \end{tabular}
    \end{minipage}    
    \caption{The left table presents the inference time of different models on the OPV2V dataset, while the right table shows the inference time of different components in mmCooper.}
    \label{tab:inference time}
\end{table*}

\begin{table}[h]
    \centering
    \caption{Comparison of detection performance when using DBA or cross attention in the BBox Filtering \& Calibration (BFC) module in AP@0.7/0.5 on OPV2V, DAIR-V2X, and V2XSet respectively.}
    \centering
    \scalebox{1.0}{
    \begin{tabularx}{\textwidth}{
         >{\centering\arraybackslash}c|
         >{\centering\arraybackslash}c|
         >{\centering\arraybackslash}c|
         >{\centering\arraybackslash}c}
         \cline{1-4}
        \textbf{Module}&\textbf{OPV2V}&\textbf{DAIR-V2X}&\textbf{V2XSet}\\
         \cline{1-4} 
         Cross Attn& 74.04/88.86 & 47.69/63.15   & 64.42/83.71 \\
         \cline{1-4}
         DBA& \textbf{78.11}/\textbf{88.93} & \textbf{48.27}/\textbf{65.12} & \textbf{65.86}/\textbf{84.40}\\
         \cline{1-4}         
    \end{tabularx}
    }
    \label{tab:DBA}
    \vspace{-18pt}
\end{table}

\subsection{Impact of Varying the Number of Agents}
\label{sec:number of agents}
\begin{figure*}
    \begin{center}
    \includegraphics[width=0.7\linewidth]{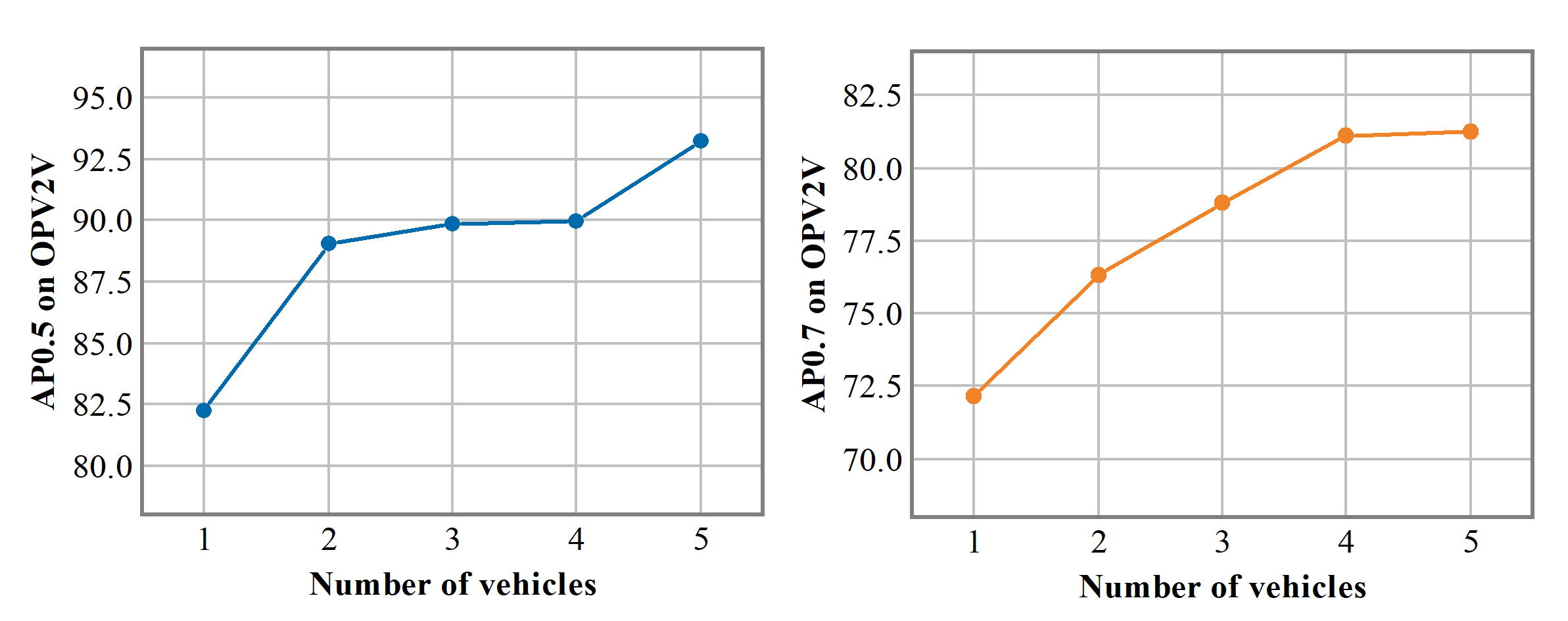}
    \vspace{-10pt}
    \caption{Impact of Varying the Number of Vehicle.}
    \label{fig:vehicle num}     
    \vspace{-18pt}
    \end{center}
\end{figure*}

\cref{fig:vehicle num} illustrates the impact of varying the number of agents on detection performance in the OPV2V dataset. It can be observed that as the number of collaborating agents increases, the detection performance also improves.

\section{Additional Qualitative Evaluation Results}
\label{sec:add_qualitative}
\subsection{Visualization of Detection Result}
\label{subsec:det}
\begin{figure*}
    \includegraphics[width=\linewidth]{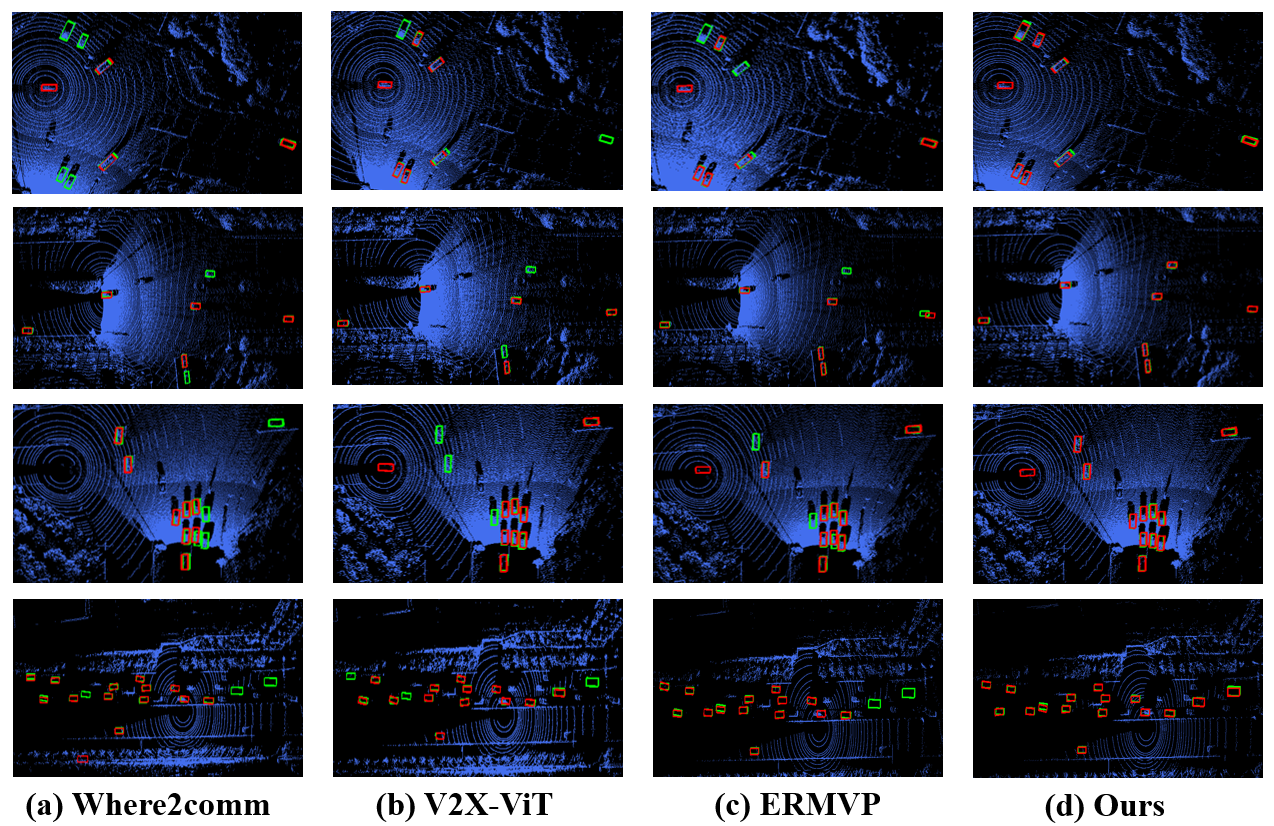}
    \vspace{-10pt}
    \caption{More visualization comparison of detection results on the DAIR-V2X dataset. Green and red boxes represent the ground truth and the model-predicted bounding boxes, respectively.}
    \label{fig:more_vis}     
    \vspace{-18pt}
\end{figure*}

We provide additional qualitative results on DAIR-V2X datasets. \cref{fig:more_vis} present visualizations of different road scenarios. Our proposed mmCooper method can detect almost all ground truths without any false positives. The results demonstrate that our proposed mmCooper achieves outstanding detection performance across various scenarios.

\subsection{Visualization of Multi-stage Fusion}
\label{subsec:multi fusion}
\begin{figure*}
    \includegraphics[width=1\linewidth]{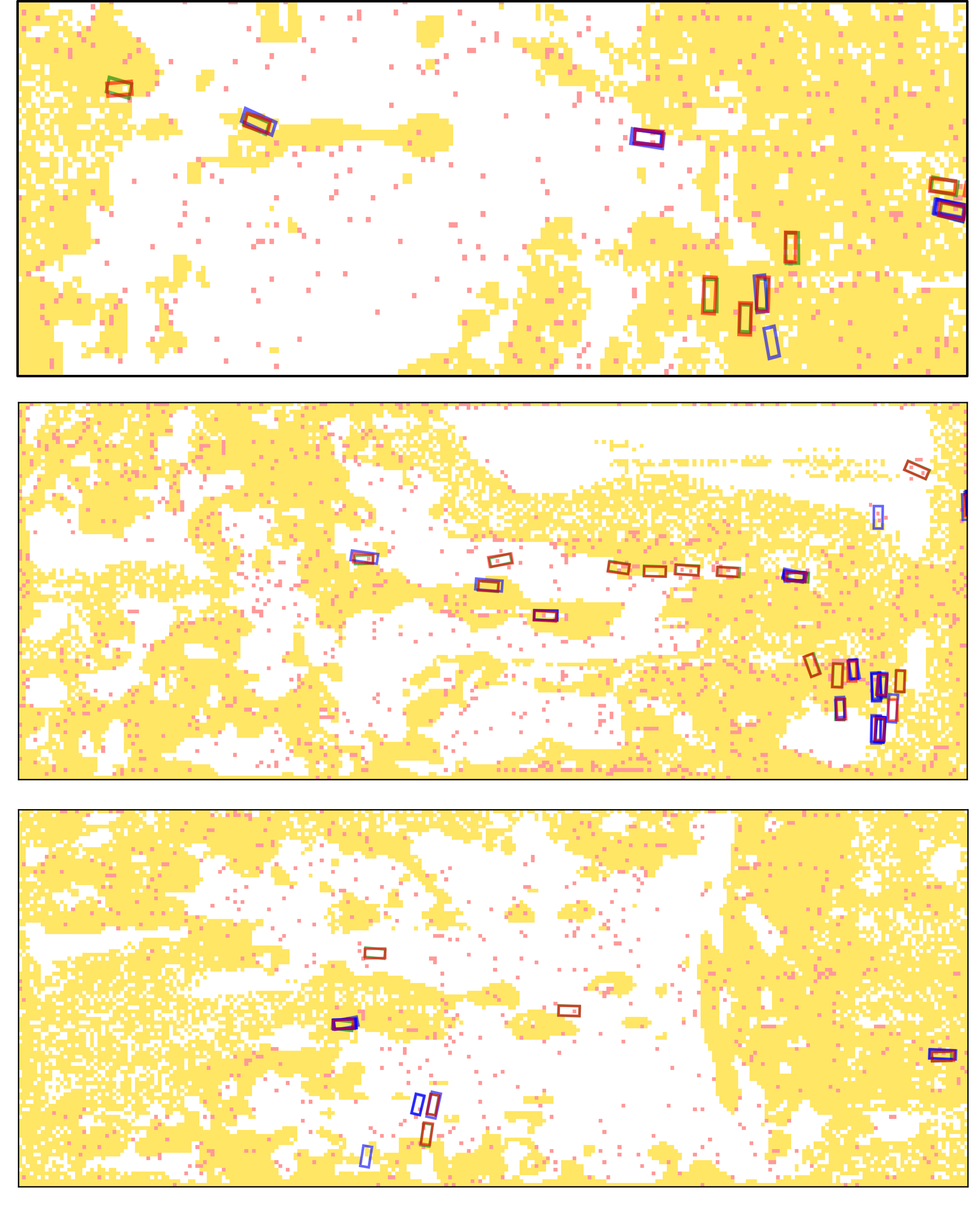}
    \vspace{-10pt}
    \caption{Visualization of two-stage fusion on DAIR-V2X. We show the results from the Confidence-based Filtering Generation Module, including discarded information (white background), BBoxes for transmission (red dots), and features for transmission (yellow background). We also show the BFC module results, including uncalibrated BBoxes (blue boxes), calibrated BBoxes (red boxes), and ground truth BBoxes (green boxes).}
    \label{fig:more_vis_stage}     
    \vspace{-18pt}
\end{figure*}

We provide more visualizations for multi-stage fusion in different scenarios. As shown in the \cref{fig:more_vis_stage}, our mmCooper effectively performs dynamic allocation of intermediate-stage features and late-stage bounding boxes across different scenarios while also refining the received bounding boxes.

